\documentclass[lettersize,journal]{IEEEtran}
\usepackage[utf8]{inputenc}
\usepackage[T1]{fontenc}
\usepackage{amsmath,amsfonts}
\usepackage{algorithmic}
\usepackage{algorithm}
\usepackage{array}
\usepackage[caption=false,font=normalsize,labelfont=sf,textfont=sf]{subfig}
\usepackage{textcomp}
\usepackage{stfloats}
\usepackage{url}
\usepackage{verbatim}
\usepackage{graphicx}
\usepackage{cite}

\usepackage{multibib}

\usepackage{xparse}
\newcites{review}{GPT Game Literature}

\def\BibTeX{{\rm B\kern-.05em{\sc i\kern-.025em b}\kern-.08em
    T\kern-.1667em\lower.7ex\hbox{E}\kern-.125emX}}
\begin{document}

\title{GPT for Games: An Updated Scoping Review (2020-2024)
}

\author{Daijin Yang, Erica Kleinman, and Casper Harteveld
\thanks{Daijin Yang, Erica Kleinman, and Casper Harteveld are with the College of Arts, Media, and Design, Northeastern University, Boston, MA 02115 USA}
\thanks{Manuscript received November 30, 2024; revised February 28, 2025.}
}


\markboth{Journal of \LaTeX\ Class Files,~Vol.~14, No.~8, August~2021}%
{Shell \MakeLowercase{\textit{et al.}}: A Sample Article Using IEEEtran.cls for IEEE Journals}
\IEEEpubid{0000--0000/00\$00.00~\copyright~2021 IEEE}

\maketitle

\begin{abstract}
 

Due to GPT's impressive generative capabilities, its applications in games are expanding rapidly. To offer researchers a comprehensive understanding of the current applications and identify both emerging trends and unexplored areas, this paper introduces an updated scoping review of 177 articles, 122 of which were published in 2024, to explore GPT's potential for games. By coding and synthesizing the papers, we identify five prominent applications of GPT in current game research: procedural content generation, mixed-initiative game design, mixed-initiative gameplay, playing games, and game user research. Drawing on insights from these application areas and emerging research, we propose future studies should focus on expanding the technical boundaries of the GPT models and exploring the complex interaction dynamics between them and users. This review aims to illustrate the state of the art in innovative GPT applications in games, offering a foundation to enrich game development and enhance player experiences through cutting-edge AI innovations.

\end{abstract}


\section{Introduction}

\IEEEPARstart{T}{he} advent of large language models (LLMs) presents new opportunities for games. Leveraging their powerful natural language processing and generative capabilities~\cite{GPT4}, LLMs have demonstrated remarkable versatility across various applications~\cite{GPTsurvey}, including information extraction~\cite{entityextraction}, question-answering~\cite{QA1,QA2}, text generation~\cite{textgeneration}, programming tasks~\cite{programming}, and creativity support~\cite{aiasactive,creativitysupport}. This versatility underscores their potential to impact various aspects of games~\cite{GPT4Game}, where many tasks --- such as programming, story writing, and game-player interactions --- involve a large amount of text-based work. Moreover, the unique generative capabilities of LLMs present distinct opportunities for both game design and gameplay, enabling them to move beyond pre-set, mechanical content and introduce innovative, dynamic elements that bring fresh ideas to both game designers and players.

Among all LLM models, the Generative Pre-trained Transformer (GPT) series has seen the most frequent and widespread applications in games~\cite{GPT4Game}. Given GPT's strong language processing capabilities~\cite{GPT3,GPT4}, it has been extensively used to generate and manage text-based game content, such as character dialogue~\citereview{14} and entire stories~\citereview{7}, and employed for creative purposes, such as generating ideas for board games~\citereview{37} or assisting with programming~\citereview{93,102}. 

As the use of GPT in games continues to expand rapidly~\cite{GPT4Game}, reviewing emerging cases can help researchers differentiate between recent developments and earlier work, allowing for a more comprehensive understanding of the current state of GPT in gaming and highlighting key trends and gaps. While existing reviews have explored the application of LLMs in games, they often lack systematic searches of published studies~\cite{review1}, focus on a single application area~\cite{review3,review4}, or miss recent research~\cite{review2}. Our previous review~\cite{GPT4Game} of 55 GPT-for-games papers included a systematic search of mainstream databases in the years 2021-2023, but did not capture the explosion of recent studies in 2024. Thus, we present an updated review that incorporates the latest research to provide researchers with a perspective on GPT for games from 2020 to 2024.

In this work, we extend our previous review~\cite{GPT4Game} using the same methodology. We identified 122 new articles that were published in 2024 and incorporated them, resulting in 177 papers. Open coding of the new papers did not identify any new use case categories. Consequently, we assigned these papers to the existing five categories: procedural content generation, mixed-initiative game design, mixed-initiative gameplay, playing games, and game user research. For each category, we present relevant examples, highlight the trends and differences between recent and earlier studies, and suggest new directions.

\IEEEpubidadjcol
\section{Methodology}
In our review, we used the keywords ``game'' and ``GPT'' to search for relevant articles across the following databases: ACM Digital Library, IEEE Xplore, Springer, and AAAI. 

These databases were selected due to their reputation in CS and AI research, offering access to papers focused on games and GPT from a technical perspective, considering that early AI research in games is inherently technical~\cite{ACM,IEEE}. For ACM, IEEE, and Springer, we performed searches directly through the search engines on their respective websites. However, for AAAI papers, since the AAAI online library lacks an advanced search function, we used Google Scholar as an alternative, restricting the results to articles from aaai.org. In all cases, we conducted full-text searches within the databases.
The search date was January 20, 2025.

\begin{figure}[htpb]
\begin{minipage}[b]{1\linewidth}
  \centering
  \centerline{\includegraphics[width=8cm]{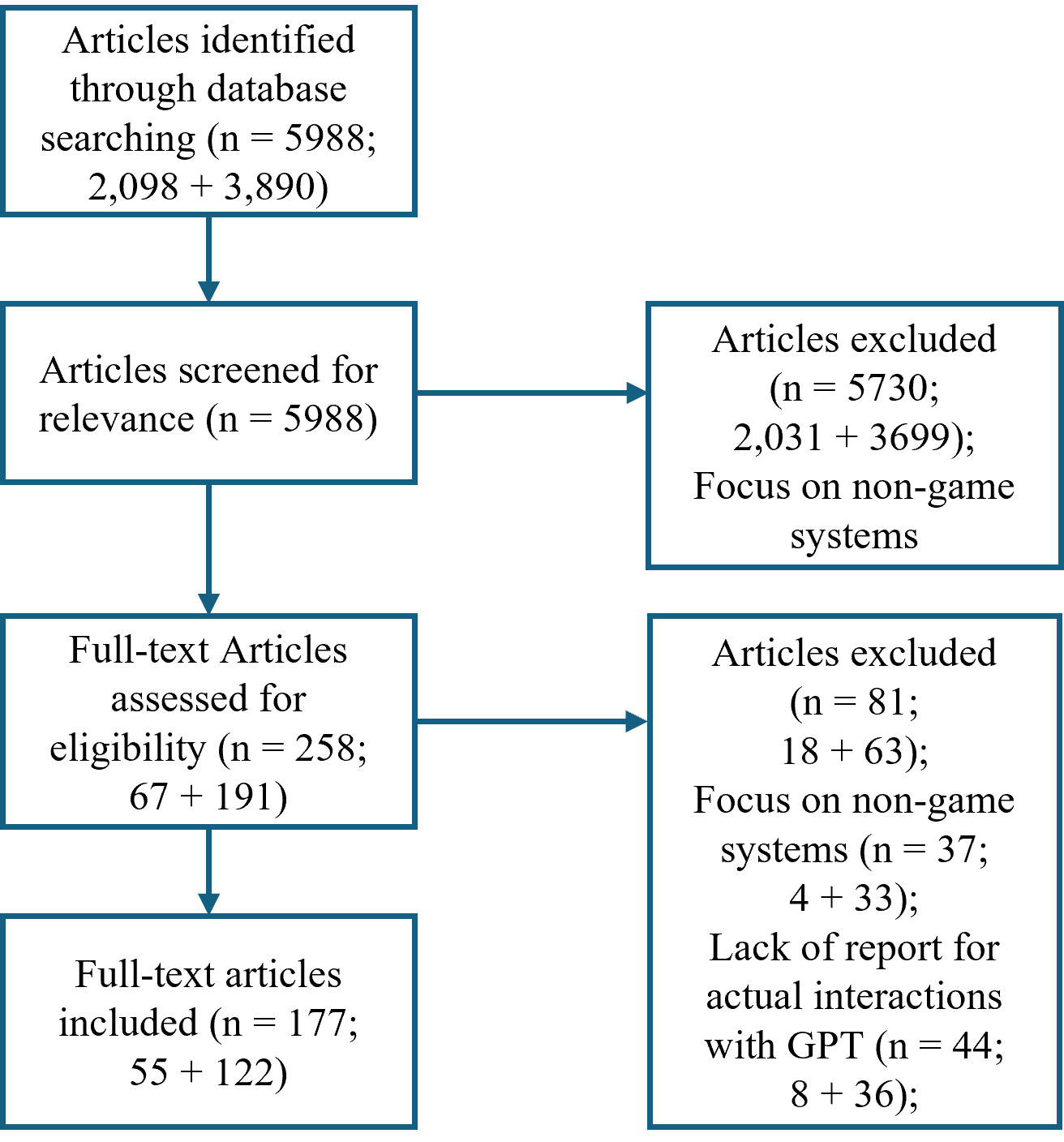}}
\end{minipage}
\caption{Flow Diagram for Systematic Search and Screening. The number before the plus sign shows previously identified articles, and the number after shows newly identified ones.}
\label{fig:screen}
\end{figure}

As shown in Figure~\ref{fig:screen}, the preliminary search using the aforementioned keywords resulted in 5,988 papers (3,890 new papers from 2024, referred to as np.), including 2,453 (1,576 np.) articles from ACM, 2,472 (1,569 np.) from IEEE Explore, 144 (82 np.) from AAAI, and 919 (663 np.) from Springer. We then screened the preliminary search results based on the following criteria:
\begin{itemize}
    \item [1)] The articles must involve the study of a system related to digital or analog games.
    \item [2)] The articles must report interactions of either authors or their participants with any version of GPT.
    \item [3)]The articles were published before  January 1, 2025.
\end{itemize}

After reviewing the titles and abstracts, 258 (191 np.) papers directly related to games and mentioning GPT advanced to the next stage. A large number of articles were excluded at this stage because `game' is a broadly used term, not solely referring to the context of games discussed in this paper, which we define as a structured activity with rules, goals, and challenges, where players interact to achieve objectives, often for entertainment or skill development. Upon a detailed examination of the 258 papers, 37 (33 np.) articles were excluded based on the first criterion, and 44 (36 np.) articles were excluded based on the second criterion.

Ultimately, 177 articles were included in the final literature synthesis. Following the methods from our previous review, we tracked the trends by recording the publication dates and the primary GPT models presented in the articles. If an article used multiple models, we counted the primary one, or the most recent if they had equal roles. 
ChatGPT is regarded as a separate model because it can operate on different versions of GPT, including GPT-3.5 and GPT-4, and is accessible only through web-based interaction. Additionally, most studies that used ChatGPT do not specify the exact version used. More specific versions of GPT (e.g., GPT-3.5-turbo-1106, a variant of GPT-3.5) are not tracked because most studies did not specify the exact version they used.

In our previous review, one researcher open-coded~\cite{opencoding} the papers to identify content generated by GPT, its use in game-related systems, and user interactions. Two additional researchers reviewed and refined the categorizations. This process revealed five major use case categories: procedural content generation (PCG), mixed-initiative game design and development (MIGDD), mixed-initiative gameplay (MIG), playing games (PG), and game user research (GUR). Each category was further divided into subcategories based on the type and usage of the generated content. For the new papers, we followed the same coding procedure, and no new major categories emerged. However, subcategories in each major use case were updated based on new interactions and designs that emerged in the new papers. 

\section{Results}

\subsection{General GPT Usage Trends of GPT for Games}

\begin{figure}[htpb]
\begin{minipage}[b]{1.0\linewidth}
  \centering
  \centerline{\includegraphics[width=9cm]{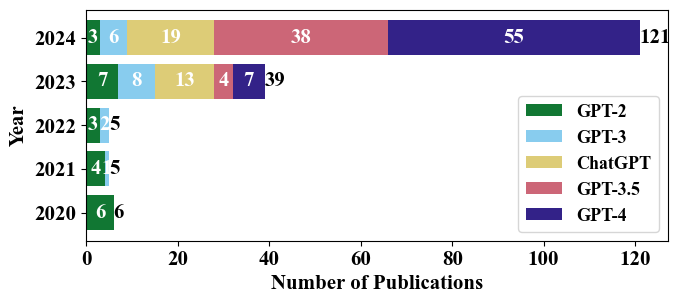}}
\end{minipage}
\caption{The General GPT Usage Trends of GPT for Games. 122 new studies were published in 2024, up from 39 in 2023, showing a significant growth trend compared to previous years. There was also rising interest in the latest GPT models like GPT-3.5 and GPT-4. One paper~\cite{142} did not mention the GPT version it used.}
\label{fig:trend}
\end{figure}

As shown in Figure~\ref{fig:trend}, in 2024, there have been 122 new studies, reflecting a huge increase compared to 39 studies in 2023. Additionally, the interest in new models, such as GPT-3.5 and GPT-4, has also grown. This sharp rise aligns with the performance improvements of GPT models~\cite{gpt2vsgpt3}, suggesting that studies will explore the application of the newest GPT models in games in the future. 

\subsection{General Research Trends of GPT for Games}
\begin{figure}[htpb]
\begin{minipage}[b]{1.0\linewidth}
  \centering
  \centerline{\includegraphics[width=9cm]{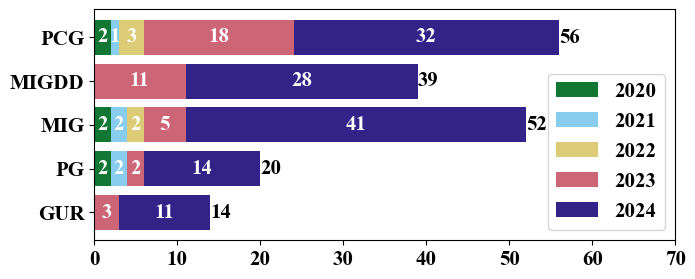}}
\end{minipage}
\caption{The General Trends of Each Category. In 2024, most papers focused on PCG and MIG, with growing interest in MIGDD, PG, and GUR. Three papers were counted in both PCG and MIG. One paper was counted in both PCG and GUR. In all, 56 papers focused on PCG, 39 on MIGDD, 52 on MIG, 20 on GPT playing games, and 14 on GUR.}
\label{fig:trendyear}
\end{figure}

The number of papers that fell into each category can be seen in Figure~\ref{fig:trendyear}. Due to the limitations of GPT-2’s generative capabilities, early studies (2020–2022) primarily focused on improving its output quality~\cite{GPT4Game}. Discussions at that time centered on whether GPT-2 could generate coherent, grammatically correct text within a game context compared to other models or whether it could perform specific non-text tasks, such as level generation. These efforts typically involved fine-tuning GPT-2, directly modifying its architecture to better suit gaming applications, or integrating it with other techniques, such as knowledge graphs, to enhance its output quality. These early explorations set the boundaries for GPT’s application in games.

By 2023, with the release of GPT-3 and subsequent models offering significant improvements in output quality~\cite{GPT3,GPTsurvey}, research shifted away from evaluating GPT’s text generation capabilities. Instead, studies increasingly described various game systems built around GPT and examined how these systems differed from non-GPT-based alternatives. Notably, research in 2023 rarely discussed complex technical methods to improve GPT-3’s output. However, some papers still reported issues such as incoherence or hallucinations in GPT-generated content~\citereview{35,10}.

In 2024, compared to the previous year, a growing number of studies explored solutions to address the instability of GPT-generated outputs in gaming applications rather than merely introducing new use cases (or revisiting GPT-2-based cases) and reporting their limitations. While the trend remains emergent, studies have begun leveraging advanced prompt engineering techniques to refine model outputs~\citereview{122,147}. In addition to prompt engineering, techniques such as Retrieval-Augmented Generation (RAG)~\cite{rag} and fine-tuning were also explored with newer GPT models to enhance their output based on game context.
Studies also have experimented with multi-GPT frameworks, enabling models to handle more complex, multi-task scenarios rather than focusing on isolated tasks~\citereview{149,98}. This shift may indicate an evolving trend.

Beyond these developments, we observed that only a limited number of 2024 studies explored user experience aspects. Similarly, few studies have utilized the multimodal  capabilities of newer models.
In the following sections, we will detail five key applications of GPT in gaming—PCG, MIGDD, MIG, PG, and GUR—highlighting new findings from 2024 and how they differ from earlier works.

\subsection{Use Cases of GPT for Games}
In this section, we report on the different use case categories that emerged from our analysis in detail.

\subsubsection{GPT for Procedural Content Generation (PCG)}
\label{para:pcg}

In this use case, GPT generates game content during gameplay according to constraints defined by the designers during development~\cite{pcgsurvey}.

\label{section:storygeneration}
\textbf{GPT for textual content generation (\textit{n} = 25,~\citereview{ 34,38,15,46,7,23,28,22,14,54,20,25,26,57,58,61,63,69,86,92,123,91,159,135,163}).} 
In this category, the majority of the papers (n = 24) focused on generating stories.
In 2024, nine studies focused on story generation, all utilizing the latest GPT models (either GPT-4 or GPT-3.5). Compared to previous work, these new studies demonstrate a stronger connection between GPT-generated stories and gameplay. Rather than merely focusing on language style or world-building, the models now incorporate additional game-related information, such as changes in world state or player input, allowing the narrative to adapt to the player’s current gameplay context. For example, in~\citereview{58}, the authors implemented a generative broadcasting system using GPT-3.5. In this system, GPT generates news reports, stories, ads, and interviews based on the current game state (events and player actions) as well as a general knowledge base (game setting, characters, and story background). 

This high adaptability of text generation presents additional challenges for GPT models, as they must process and respond to real-time game content while generating consistent output. To address these challenges, these studies often integrate GPT with other technologies. For instance, in~\citereview{91}, the authors leveraged various modules from the LangChain~\cite{langchain} framework to enhance GPT-4's capabilities in retrieving career data, generating career plans, and tracking interaction history. 

Research that enhances GPT's text generation by integrating it with other technologies draws inspiration from earlier studies using GPT-2. All studies (\textit{n = 5}) involving GPT-2 needed fine-tuning to better align the model’s output with specific contexts, which demanded a dataset closely related to the research goals~\citereview{46}. For instance, in~\citereview{26}, GPT-2 was fine-tuned using fantasy stories to create narratives for a Dungeons and Dragons (D\&D) game. To address GPT-2's limitations in language processing and memory capacity, additional techniques were employed, such as~\citereview{7} using a knowledge graph alongside GPT to generate more consistent stories. 
The authors first use another AI model called AskBERT to extract a knowledge graph consisting of locations, characters, and objects. Then, GPT-2 is used to generate consistent descriptions of these entities based on the relationships in the graph.
Additionally, in~\citereview{26}, BertScore~\cite{bertscore} was applied to evaluate GPT-2's outputs, ensuring that only high-quality texts were presented to players, mitigating issues with low-quality content.
Just as knowledge graphs were used in~\citereview{92} to improve GPT-4's output, 
These techniques are not limited to enhancing GPT-2's performance --- they also hold potential for improving the capabilities of the latest models.

Besides story generation, two papers used GPT to generate other textual content. In~\citereview{135}, GPT was asked to generate quiz distractors at runtime in an educational game while students played, as well as personalized review exercises, reducing teacher workload and adapting content based on student performance. In~\citereview{163}, the authors explored the use of Large Language Models (LLMs) to generate Connections puzzles, a word association game by The New York Times, using a Tree of Thoughts prompting approach~\cite{treeofthoughts}. A user study found that AI-generated puzzles were competitive with human-made ones in creativity and difficulty, highlighting LLMs' potential to generate word connection puzzles without the help of humans.

\textbf{GPT for quest generation (\textit{n} = 4,~\citereview{27,41,47,31}).}
Unlike a story, a quest involves a series of actions the player must complete~\cite{quest}. GPT-2 was the primary model used for quest generation. Common approaches across these studies involved fine-tuning the GPT-2 model on annotated datasets specific to RPG quests and applying prompt engineering to guide the model's output towards objectives and dialogues that align with the structure and style of existing game content. For instance, in~\citereview{27}, the authors experimented with GPT-2 to generate quests within a ``World of Warcraft'' setting. They fine-tuned the model using a quest text corpus, enabling it to produce quests that not only aligned with the game's themes but also retained logical consistency and creativity. Additionally, in~\citereview{41}, researchers explored integrating knowledge graphs for quest generation, allowing the model to incorporate game-world elements such as locations, NPCs, and items into the quests. This method ensured quests were contextually relevant and personalized, enhancing player immersion by tailoring content to the game's lore and the player's actions.
 
 \textbf{GPT for level generation (\textit{n} = 6,~\citereview{45,36,43,56,122,132}).}
In recent research, published in 2024, the latest GPT models (GPT-3.5, ChatGPT, and GPT-4) have been used to generate levels using only prompt engineering. For example, in~\citereview{122}, the authors discuss how different prompting methods affect the quality of level generation in Angry Birds~\cite{angrybirds}, a physics-based game where players launch birds to knock down pigs. They found that the data store and retrieval method, which involves pre-storing level components and allowing GPT to retrieve the appropriate ones, resulted in the most stable and high-quality levels. Beyond 2D level generation,~\citereview{56} introduces a method for generating 3D Minecraft~\cite{minecraft} buildings, which could potentially be adapted for 3D level generation. GPT-4 converts a simple user description (such as specifying only building materials) into a detailed description that includes the size, structure, and materials of the building. It then transforms this detailed description into a JSON file, which is used by Python to build the structure in Minecraft. The model also includes a repair module to fix errors, ensuring accurate generation of elements like walls and doors. However, the integration of 3D level generation with gameplay still requires further exploration.

\textbf{GPT for character generation (\textit{n} = 18,~\citereview{1,21,25,101,130,106,73,88,89,116,127,99,105,121,104,133,145,165}).} 
In 2024, the use of GPT to generate in-game characters has attracted significant research attention (\textit{n = 15}). Unlike previous approaches that focused only on generating character descriptions and settings, most recent work (\textit{n = 12}) used GPT to drive characters in games, enabling them to act in ways consistent with their personality and backstory. For example, in~\citereview{106}, the authors assigned each NPC a backstory, personality, and a series of ordered goals and conditions. GPT-4 then used this information to generate dialogues for these characters. 

Other techniques were introduced to preserve the characters' personalities better and prevent hallucinations over extended gameplay. In~\citereview{127}, for instance, the authors implemented a behavior tree, ensuring GPT only generates dialogue when a player triggers specific actions, thus maintaining control over the output. In~\citereview{101}, a unique memory storage mechanism was designed, where GPT summarizes recent conversations (stored in a short-term memory), transfers a condensed version to a long-term memory, and then removes the recent dialogue. When generating new dialogue, GPT retrieves data from both memory storage, producing responses that reflect its accumulated experiences. This human-like memory behavior helps ensure continuity in long-term interactions, allowing GPT to generate more human-like responses.
Besides generating dialogue, GPT can also assist in character animation. For example, in~\cite{130}, the authors use GPT-3.5 to generate facial expression descriptions based on the Facial Action Coding System~\cite{facs} (FACS) and body movement descriptions based on Laban Movement Analysis~\cite{LMA} (LMA), considering the dialogue and the character's personality.

\textbf{GPT for recognizing user input (\textit{n} = 2,~\citereview{142,167}).}
In 2024, we found two papers that utilized GPT to transform user inputs into texts that can be further processed by the game systems. In~\citereview{142}, the authors employed GPT to recognize and categorize players' emotions in their text inputs. The outputs of GPT were then used to generate different colors on the screen.
In~\citereview{167}, the authors used the multimodal abilities of the latest model, GPT-4V, to recognize players' hand gestures, which were captured by a camera in the form of video. The game reacted to different players' gestures accordingly.

\textbf{GPT for other PCG uses (\textit{n} = 2,~\citereview{17,5}).}
Our previous review revealed two other instances of GPT being used to generate content for games. In the first case~\citereview{17}, GPT was employed for music generation. The authors introduced Bardo Composer, a system that creates background music for tabletop role-playing games. It utilizes a speech recognition system to convert player dialogue into text, which is then classified based on an emotional model. The system generates music that conveys the desired emotion using a novel Stochastic Bi-Objective Beam Search algorithm~\cite{SBBS}.
In the second case~\citereview{5}, GPT was used to generate real-time commentary. The authors developed a prompt engineering approach with GPT-3.5 to produce dynamic commentary for fighting games. Their work highlighted the significant impact of prompt design on the quality of the generated commentary, with users showing a preference for simpler prompts to create more engaging experiences.

\subsubsection{GPT for Mixed-Initiative Game Design and Development}
Similar to the previous category, GPT was employed to generate content for design and development purposes. However, unlike the previous category, the content generation followed an iterative process, where the designer collaborated with GPT through multiple rounds to create the final content, which was later incorporated into the game~\cite{mixedinitiativeinterface}.

\textbf{GPT assistance in creating game scenarios (\textit{n} = 23,~\citereview{39,30, 49,50,40,107,110,97,109,98,67,75,78,94,103,134,168,169,173,136,149,176,164}).} 
The most direct application uses GPT to assist with writing scenarios for game stories. 
A simple example can be found in~\citereview{94}, where the authors used GPT-3.5 to generate stories directly and deliberately retained certain hallucinations (incorrect information) produced by the language model. This was done as part of a game designed to investigate players' perceptions of deceptive behavior. 
Another noteworthy example is in~\citereview{103}, where the authors designed two distinct story creation modes for scriptwriters with varying levels of experience. For less experienced writers, GPT offered more extensive support, allowing them to input only minimal story details as a starting point, while GPT actively generated elements such as characters and scenes. For more experienced writers, GPT played a more passive role, providing summaries and assistance without interfering with their creative process. This demonstrates that GPT's support can be adjusted to better serve different user needs in game design~\cite{dynamicscaffold}.

In addition to a simple collaboration flow between humans and AI agents, in~\citereview{149}, the authors presented an LLM architecture that could assist in designing a multi-modal digital narrative game. This system utilized a hierarchical network of GPT agents where multiple agent teams collaboratively decomposed and structured the narrative game into several components, including settings, story beats, scenes, screenplays, and characters. Each team consisted of expert agents, who created content, and critic agents, who refined it through iterative feedback. Finally, a compose agent assembled all components into a cohesive, long-duration interactive digital narrative game. This process allowed both AI-driven automation and human intervention, ensuring an adaptive and high-quality storytelling experience.

\textbf{GPT assistance in designing game mechanics and rules (\textit{n} = 8,~\citereview{2,33,11,37,12,76,124,151}).} 
Three studies published in 2024 used GPT to assist in designing game mechanics and rules~\citereview{76,124,151}. In~\citereview{76}, the authors applied few-shot prompting (where a few examples are included in the prompt) to guide GPT-4 in generating a game description using VGDL~\cite{vgdl}, a game description language framework for both rules and levels.  

In~\citereview{151}, compared to~\citereview{76}, the authors presented a more advanced framework for generating grammatically accurate VGDL descriptions from natural language using GPT through a two-stage process: Rule Decoding, which generates a minimal set of grammar rules, and Game Description Decoding, which uses these rules to iteratively refine the game description for correctness. Experimental results demonstrated that this approach significantly improved compilability, functionality, and syntactic accuracy compared to baseline prompting methods. However, recent research has not yet explored combining GPT with other technologies or design frameworks to generate game mechanics, which we discovered in our previous review, in papers such as~\citereview{11}, which used an evolutionary algorithm with GPT to create playable board games.
In~\citereview{124}, ChatGPT was employed to suggest additional mechanics for simple games. For example, GPT added teleportation and time manipulation mechanics to a space shooter game that originally featured only movement and shooting. GPT was also used to assist in programming these mechanics. The authors further conducted a user study of the improved games and found that GPT's suggestions enhanced the playability of simple games and inspired human designers.

\textbf{GPT assistance with programming tasks (\textit{n} = 11,~\citereview{12,51,98,124,79,59,93,102,65,128,172}).} 
\label{section:programming}
Some studies~\citereview{12,98,124} have integrated programming support directly into tools for scene creation or game mechanics development. For instance, the research discussed in previous sections on 3D scene creation~\citereview{98} and suggestions for basic game mechanics~\citereview{124} included code implementation and programming assistance within their systems.
Regarding programming assistance,~\citereview{93} provides a detailed account of the opportunities and challenges of using GPT for programming support in VR development. They found that GPT can significantly assist novice developers by offering guidance on coding, troubleshooting, and feature implementation, helping them overcome common challenges in Unity-based VR development. However, for more complex VR tasks, such as handling interactive elements and real-time physics, GPT may provide inaccurate advice or incomplete responses, requiring additional user oversight.

Beyond direct coding support, two examples are worth mentioning~\citereview{79,65,128}. In~\citereview{79}, the authors introduced GlitchBench, a new benchmark designed to test the ability of multimodal LLMs to detect glitches in video games. The authors found that GPT-4 achieved the highest accuracy, although it could correctly detect only 43.4\% of the glitches. In the future, enhancing the model’s detection capabilities could facilitate its use in game quality testing and debug programming.
In~\citereview{65}, the authors used ChatGPT to generate reward functions for a deep reinforcement learning (DRL) algorithm~\cite{drl} based on in-game information, assisting in the training of DRL agents. ChatGPT first generates initial reward functions based on textual descriptions of the game and environmental variables. The DRL agents then interact with the game environment using these reward functions, gradually optimizing their strategies.

\subsubsection{GPT for Mixed-initiative Gameplay}
Unlike the previous section, which leverages GPT to aid in design, this section focuses on GPT as an aid during play.
Compared to previous years, 2024 has seen a significant increase in attention to this category, with 41 new papers published.

\textbf{GPT to aid narrative co-creation in games (\textit{n} = 29,~\citereview{9,44,16,19,48,35,24,8,126,74,115,118,120,62,64,66,72,100,81,138,140,158,144,141,146,155,177,150,18}).}
In this category, most papers focused on story co-creation, which involves players and GPT taking turns to contribute to a story, adding sentences or paragraphs to the narrative. The process typically begins with a prompt, either provided by GPT or the players, to start the story. GPT then generates new content by using the existing story as input for its next contribution. 
For example, in~\citereview{118}, the authors designed Snake Story, a story co-creation game where players use the navigation and eating mechanics of the classic Snake game to decide which GPT generated sentences to add to the story.

In 2024, new research has begun to move beyond simply co-creating stories with GPT to address more serious topics, including climate change~\citereview{74}, cultural heritage protection~\citereview{155}, medicine~\citereview{138}, AI education~\citereview{81,158}, and other educational topics~\citereview{141,144,100}.
For instance, in~\citereview{74}, the author developed Eternagram, a story co-creation game that assesses players' attitudes toward climate change through story co-creation. As players progress, they gradually develop a future world devastated by climate change with GPT-4. The research showed that this gamified approach could explore players' perspectives on climate issues in depth.
In the educational space,~\citereview{81} presented Hacc-Man, a game centered around the concept of ``jail-breaking'' LLMs. Players interact with GPT through a retro arcade-style computer, attempting to coax GPT into unethical or unsafe behavior, such as tricking it into revealing another patient's health information in a hospital scenario.
Overall, these studies highlight the potential of GPT-driven games to engage the public in reflections on and discussions about serious topics.

In addition to story co-creation, we identified two other instances that employed GPT to aid creating emergent narratives by enabling new mechanics based on player interactions in the game~\citereview{18,177}. In~\citereview{18}, the authors introduced the Real-time Creative Element Synthesis Framework (RCESG). This framework was demonstrated in the game Create Ice Cream, where players could freely combine words. GPT generated new words based on these combinations --- for example, merging ``fire'' and ``leaves'' to produce ``burn'' --- which were then used to tackle challenges in Story Mode, such as defending against an alien invasion or crafting a royal feast.
In addition to text-based interaction,~\citereview{177} introduced a system that implemented new mechanics into the game at runtime, enabling emergent narratives by dynamically shaping gameplay based on player actions. By evaluating prototypes and developer interviews, the study highlighted the system's potential to enhance player agency and content diversity while also raising concerns about quality control, workflow integration, and balancing unpredictability with meaningful storytelling.

\textbf{GPT for providing feedback and guidance to players in games (\textit{n} = 20,~\citereview{114,111,77,129,71,96,119,60,83,116,88,89,152,175,148,154,170,143,174,162}).}
In 2024, feedback and guidance, especially within educational games, garnered significant attention, with all papers in this subcategory published this year. For instance, in~\citereview{71}, the authors discuss the integration of GPT into game-based learning environments to offer personalized, subject-specific feedback on players' in-game actions, answer their questions during gameplay, and provide relevant explanations or demonstrations based on the game's tasks and learning content. The study shows that players in GPT-assisted environments exhibit stronger intrinsic motivation and cognitive engagement compared to those in purely game-based settings.

In addition to feedback, some papers used GPT to provide action suggestions~\citereview{129,148,154,170,143,174}. For instance, in~\citereview{174}, the authors explored how GPT can generate action suggestions for users in VR by leveraging task descriptions (overall test goal), history of completed actions, and the user's current state (position and view direction).
Another interesting example is~\citereview{129}, where the authors used GPT to simulate a player's inner monologue in a 3D creepy hotel game environment. When players triggered events, such as opening doors or inspecting objects, GPT-4 would generate insights, comments, or suggestions in the form of self-talk, aiding the player’s progression and advancing the narrative. 
 
One other unique work is particularly worth mentioning. In~\citereview{77}, the authors utilized GPT-3.5 to handle natural language commands given by players to NPCs, adjusting the NPCs’ strategic priorities accordingly. For instance, when a player says, ``protect me,'' the system increases the priority of defensive goals for the NPC.
In all, these works demonstrate the potential of GPT to transform gameplay and enhance the overall player experience.

\textbf{GPT to support game masters (\textit{n} = 3,~\citereview{10,53,70}).}
In this category, with one new study published in 2024~\citereview{70}, GPT was applied to assist Game Masters (GMs) in Dungeons \& Dragons (D\&D) or other tabletop role-playing games (TTRPGs). 
These studies explored how GPT-3 and ChatGPT can provide creative support by generating enemy descriptions and configurations, summarizing game scenarios, and brainstorming narrative elements. 
For instance, in~\citereview{10}, GPT-3 was used to simplify and summarize the descriptions of monsters from the D\&D rulebook, helping DMs quickly grasp key details. Additionally, ChatGPT was employed to assist DMs in brainstorming encounter storylines. The authors developed separate interfaces for each function and created distinct prompts to enable GPT to perform these tasks effectively.

Notably, in a recent study~\citereview{70}, the authors explored the use of ChatGPT as an independent GM for D\&D without human assistance. They configured different characteristics for GPT-based GMs, including a base model, one agent inclined to support any player requests, and another that adhered strictly to the original story settings. The results showed that both specialized agents enhanced player experience compared to the base model, highlighting new potential interactions between GPT, GMs, and TTRPG players.

\subsubsection{GPT as a Game Player}
\label{section:GP}
Works in this category explored how GPT can autonomously play games or serve as a virtual opponent. 

\textbf{GPT for playing text-based games (\textit{n} = 13,~\citereview{4,3,55,13,42,112,125,108,80,82,147,153,161}).}
In 2024, six studies explored GPT playing text-based games, focusing on logical reasoning and social cognition. For instance, in~\citereview{112}, the authors used prompt engineering to enable GPT-4 to play Codenames, a cooperative word-guessing game. In this game, one agent provides clues to help another agent guess target words. The research revealed that, while different prompts did not significantly impact GPT's reasoning or natural language understanding abilities, they did influence its play style.

In another example~\citereview{161}, GPT-4 played the Oogiri game, a Japanese creative humor game requiring unexpected and humorous responses to text or images, and struggled with Leap-of-Thought (LoT) creativity, failing to make non-sequential, associative connections necessary for humor generation, despite its strong reasoning abilities, highlighting its limitations in creative problem-solving. To address this, the authors proposed the Creative Leap-of-Thought (CLoT) paradigm, which enhances GPT-4’s LoT ability through associable instruction tuning and explorative self-refinement, enabling it to generate more creative and humorous responses by drawing parallels between seemingly unrelated concepts.
Previous research has also featured gameplay scenarios such as deducing a word from its description~\citereview{4} and compelling a defender to utter a specific word while they attempt to avoid doing so~\citereview{3}.

\textbf{GPT for playing non-text-based games (\textit{n} = 7,~\citereview{52,131,90,87,139,160,178}).}
We identified seven studies that utilized GPT for playing various non-text-based games, including 6 published in 2024~\citereview{131,90,87,139,160,178} and one published in 2020\citereview{52}.
In~\citereview{52}, GPT-2 was used to generate plausible strategic moves in the game of Go. By training on a dataset of Go game records in Smart Game Format (SGF), the model learned to mimic the strategic styles of Go champions, producing valid and competitive game strategies.

In another example~\citereview{139}, GPT-2, through the Learning Chess Blindfolded (LCB) model~\cite{lcb}, was analyzed for its attention mechanisms in chess move prediction, revealing that early layers effectively identify strong moves while later layers refine selection without deep strategic planning, highlighting its strength in pattern recognition but limitations in long-term reasoning.

Beyond board games, in~\citereview{87}, GPT was used to play Angry Birds~\cite{angrybirds}.
GPT first selects an appropriate target based on the game rules and scene information (e.g., the size, material, and coordinates of objects). This target information is then sent to a calculator to determine the slingshot's angle, enabling the bird to be launched towards the selected target. The authors noted that while the system performed well in simple scenarios, it needed improvement when dealing with more complex firing strategies.

In addition to static turn based games, GPT was also employed to play more dynamic action games.
In~\citereview{178}, GPT-4 was tasked with playing the 1993 first-person shooter Doom by interpreting game states from screenshots and generating textual commands to control the game, demonstrating basic planning and combat abilities such as opening doors, fighting enemies, and navigating paths. However, it struggled with long-term reasoning, spatial awareness, and enemy tracking, often forgetting unseen enemies, getting stuck in corners, and exhibiting erratic movement patterns. The study found that prompting strategies incorporating multi-step planning and expert consultation improved performance, though GPT-4 remained significantly weaker than reinforcement learning-based approaches, highlighting its limitations in real-time dynamic environments.

\subsubsection{GPT for Game User Research}
Even with the inclusion of six new papers from 2024, studies focusing on GPT for game user research remain scarce, with only nine papers in total addressing this category. These studies predominantly utilize newer models, with GPT-2 not used in this category.

\textbf{GPT for processing game reviews (\textit{n = 4},~\citereview{84,29,6,171}).} 
This subcategory contains two new research from 2024~\citereview{84,171}. In~\citereview{84}, the authors used ChatGPT to generate recommended reviews for the Ant Forest game, a mobile game that rewards eco-friendly actions with virtual energy, which is then used to plant real trees in reforestation projects. The study found that recommendations generated by ChatGPT were perceived to have higher content quality, and users tended to prefer AI-generated recommendations over human-generated ones.

In~\citereview{171}, the authors presented a novel dataset annotated by experts for aspect-based suggestion mining in game reviews, capturing both explicit and implicit suggestions. Evaluated with deep learning models, including GPT-3, the dataset improves suggestion extraction. Fine-tuned on 50 examples per category, GPT-3 outperformed the SemEval 2019 dataset~\cite{semeval} in binary classification, demonstrating strong suggestion mining capabilities.

From the previous review, in~\cite{29}, the authors explored GPT-3's potential to analyze game reviews to enhance game design, focusing on how AI can provide insights into player experiences and preferences by prompting GPT-3 to answer questions from the Player Experience Inventory (PXI)\cite{pxi} based on players' game reviews. In another example from the previous review, \cite{6}, the authors investigated the use of ChatGPT for categorizing audience comments during live game streaming to increase engagement and stream value. ChatGPT was prompted to classify audience members into five predefined categories based on their comments and the live streamer's current game status.

\textbf{GPT for analyzing player behavior (\textit{n = 4},~\citereview{68,85,106,137}).} 
All three studies in this subcategory were from 2024 and feature studies in which GPT was used to examine and assess how players interacted with a game. In~\citereview{68}, GPT was employed to assess whether players' actions in a TTRPG aligned with their character settings. The authors used controlled chain of thought prompting (CCoT) for the purpose. First, GPT generated several possible behaviors based on the character's faction setting, then, GPT analyzed the player's actual in-game dialogue to check if it matched the predicted behaviors. Finally, GPT was asked to summarize its findings and develop a more general understanding of the character's faction.

In~\citereview{85}, GPT was used to predict players' emotional shifts during a coin-flipping game. The game involved four coin flips, with the likelihood of winning or losing becoming clearer as the game progressed, causing emotional changes. The study found that GPT performed well in reasoning about the direction of emotional changes (positive or negative) but struggled with predicting the intensity of emotions and coping behaviors.

In~\citereview{106}, the authors used GPT to summarize interaction logs from a story co-writing game. Based on these summaries, they built narrative graphs to describe player behavior in the game. By comparing these graphs with the original game narrative, they could identify emerging nodes—new, unplanned strategies or interactions introduced by the players. While these studies demonstrate GPT's potential to analyze and interpret player behavior in unique ways, they also highlight its limitations in fully capturing emotional intensity and nuanced player strategies.

In~\citereview{137}, the authors used GPT-3.5 to analyze human gameplay in open-world games, specifically to extract game elements from players' think-aloud transcripts and categorize them based on the knowledge used in decision-making. The authors then compared the GPT-3.5-generated list to a manually generated list, validating the identified game elements and identifying how players solve problems, interact with game elements, and use information.

\textbf{GPT for recommending games for players (\textit{n = 3},~\citereview{95,156,166}).} 
All three studies in this subcategory were published in 2024 and explored GPT's role in recommending games to players.
In~\citereview{95}, the authors designed the LLaRA system based on GPT-4, which, unlike traditional recommendation systems that rely solely on game IDs, integrated textual information such as game storylines and reviews. Additionally, LLaRA introduced complex user behavior patterns, like playtime, allowing it to understand user preferences better and recommend games that match their style and interests.

In~\citereview{156}, the authors explored how Large Language Models (LLMs) enhance sequential recommendation systems by introducing three approaches: LLM-based embeddings for item similarity (LLMSeqSim), fine-tuned LLMs for recommendation tasks (LLMSeqPrompt), and LLM-enhanced traditional models (LLM2Sequential). Experiments on multiple datasets, including games, show that LLMs significantly improve recommendation accuracy, diversity, and coverage, with fine-tuned GPT-3.5 outperforming other models.

\citereview{166} introduced a recommendation paradigm where users can express their preferences and intentions using natural language instructions, such as describing desired item attributes, specifying past preferences, or requesting recommendations for a particular context. Instead of relying solely on historical interactions, a language model, such as GPT-3.5, can be used to interpret these instructions and generates personalized recommendations by following user-provided descriptions. By fine-tuning the model with diverse instruction data, this approach can enhance recommendation accuracy, improve user interaction flexibility, and outperform traditional models in accommodating a wide range of user needs.

\textbf{GPT for other game user research. (\textit{n = 3},~\citereview{113,32,157}).} 
This subcategory contains two new research from 2024~\citereview{113,157}.
In~\citereview{113}, GPT-4 was used to classify toxic messages in games. The authors applied the Prompt Evolution Through Examples (PETE) method to optimize the prompt. Specifically, the system started with an initial prompt, generated variant versions, and selected the most effective one based on performance in the task. Through multiple iterations, the prompt was gradually optimized, improving the model's performance in toxic message classification.
In~\citereview{157}, 
Additionally, in one study from the previous review,~\citereview{32}, the authors investigated and compared the answers of human participants and generative AI (GPT-4 and ChatGPT) to interview questions about voice interaction across three scenarios: game-to-player, player-to-player, and player-to-game interactions.

\section{Discussion}

Here, we discuss directions for future research on GPT for games from both technical and interaction perspectives.

\subsection{Expanding the Technical Boundaries of the Models}

Compared to earlier studies, research published in 2024 has leveraged the latest GPT models, achieving more promising results across the five distinct use cases. However, most research still relies on basic prompt engineering to interact with the models. We believe there remains significant potential to expand the technical boundaries of these models for game applications. Here, we discuss three possible directions.

\textbf{Experiment with Broader Range of Games.} First, we encourage research to explore the technical boundaries of current GPT models by experimenting with a broader range of games, providing a foundation for potential performance improvements.

As shown in the results section~\ref{section:GP}, recent 2024 studies have expanded GPT’s capability to engage with non-text-based games beyond prior work~\cite{GPT4Game}. Beyond testing its logical reasoning in text-based games, GPT has been applied to board games like Go and chess, card games, and action games such as Doom and Overcooked, excelling in structured play but struggling with long-term planning and dynamic real-time decision-making. However, many of the games it has played are relatively constrained --- that is, they feature well-defined rules, limited possible actions, and largely predictable environments --- which does not fully challenge GPT’s ability to handle continuously evolving states, nuanced social dynamics, and real-time adaptation. For GPT’s capabilities to be truly challenged, it needs to engage with more complex, open-ended environments that require on-the-fly learning, coordination with multiple agents, and strategic thinking over extended durations.

As an example of further challenging GPT's capabilities, future research could examine GPT's performance in games with more complex rules, strategies, and states, such as real-time strategy (RTS) games~\cite{rts}. An RTS game like StarCraft II~\cite{starcraft2} or Age of Empires~\cite{ageofempires} would offer a rigorous test of GPT’s capacity for long-term resource management, tactical unit control, and high-level strategic planning—all in real-time. Beyond simply reacting to an opponent’s moves, GPT would need to anticipate enemy strategies, adapt to unexpected changes, and manage intricate economies and technology trees. In such an environment, GPT’s ability to balance short-term goals (e.g., defending a base) with longer-term objectives (e.g., securing expansions, upgrading units) becomes critical.

Additionally, a multiplayer environment --- particularly one that involves coordination and communication with multiple human or AI teammates --- would challenge GPT’s social reasoning and collaborative problem-solving skills. Examples might include cooperative survival games (Don’t Starve Together~\cite{dontstarvetogether}), large-scale MMOs~\cite{ffxiv2010}, or team-based shooters with strategic depth (Overwatch~\cite{overwatch2016}). In these settings, GPT would need to interpret and respond to teammates’ actions and goals, negotiate tactics in real-time, and adjust to unpredictable human behavior. This emphasis on emergent dynamics and communication would significantly broaden GPT’s challenge beyond the relatively rigid frameworks found in many single-player or turn-based environments. As these multiplayer scenarios often rely on a combination of textual, auditory, and visual information, leveraging newer GPT-4V models --- capable of processing both text and visual inputs --- could enhance GPT’s effectiveness. By analyzing complex visual cues, such as teammate positions, enemy movements, or in-game UI elements, GPT-4V could improve situational awareness and make more informed decisions in fast-paced environments, further expanding its potential in multiplayer gaming contexts.

\textbf{Advanced Prompt Engineering and Agentic LLMs.} Second, we encourage further exploration of advanced prompt engineering and agentic LLM techniques and their effectiveness in game-related tasks.
As shown in the results, while some studies have investigated sophisticated prompt engineering for level generation~\citereview{122} and game playing~\citereview{147,161}, and the use of agentic LLMs for scene creation~\citereview{149,98}, most research continues to rely on basic single-agent interactions and standard prompt settings. Consequently, the application of these advanced techniques beyond these areas remains largely unexplored.

Prompt engineering methods such as Chain of Thought prompting~\cite{CoT}, self-reflection loops, and structured prompt templates can enhance LLM performance by improving reasoning, consistency, and coherence. For instance, Chain of Thought prompting aids in generating structured narratives and gameplay logic, while self-reflection loops iteratively refine mechanics for better player engagement. Likewise, structured prompt templates ensure consistency across various elements like quests, characters, and progression systems. Our results indicate that although some research has adopted these approaches, their effectiveness has yet to be systematically evaluated across diverse game design scenarios. Regardless, while prompt engineering is well-suited for handling static tasks with predefined structures, it remains limited in addressing complex, iterative processes that require adaptive decision-making.

Agentic LLMs --- autonomous or semi-autonomous agents powered by LLMs --- offer a promising alternative to prompt engineering. Unlike single-response systems, agentic LLMs iteratively refine their outputs, interact with external tools, and collaborate with other agents to tackle complex, multi-faceted tasks. Tasks in all five use case categories are highly complex and specialized. For example, game design and development require GPT to handle diverse components such as game mechanics, user experience, and programming~\cite{gamedesignmulti}. In other fields characterized by creativity and complex cross-disciplinary tasks --- such as brainstorming for design~\cite{Brainstorming1,Brainstorming2} or software development~\cite{chatdev,gamegpt} --- multi-agent language model systems have already been shown to enhance overall performance. In these systems, each GPT often plays a role focused on a specific capability or task within a complex framework, collaborating with other GPTs in different roles to address the full scope of the problem.
Although recent studies in 2024 have reported more advanced systems for supporting game designers~\cite{GPT4Game}, research on comprehensive multi-agent systems remains limited. We encourage future research to explore integrating various GPT models into a unified system~\cite{GPT4Game}, where each model specializes in particular tasks, potentially improving GPT's contextual understanding~\citereview{33}, increasing the relevance of generated content, and enhancing overall efficiency.

\textbf{Open-Weight Language Models.} Finally, we encourage researchers to explore smaller open-weight language models. As discussed in Section~\ref{section:programming}, the study presented in~\cite{128} has already demonstrated that certain smaller models can generate dialogues and stories comparable to GPT-3.5 in the gaming domain. Open-weight models offer several advantages over proprietary large models, including lower operational costs~\cite{128}, enhanced security~\cite{smallsafe}, and greater flexibility for modifications. Moreover, models that have been distilled~\cite{smallkd} or pruned~\cite{smalltrim} from larger counterparts can retain much of their performance while being more efficient. Research aimed at improving open-weight models will also contribute to the advancement of larger models, as techniques previously applied to GPT-2 --- such as fine-tuning and the integration of knowledge graphs --- have effectively enhanced newer models. In addition to providing technical insights for improving language models more broadly, open-weight models can function as modular components that enhance the output capabilities of larger systems~\cite{smallplugin}. Their adaptability allows them to be fine-tuned for specific tasks, making their outputs valuable references for broader AI applications~\cite{smallplugin}. We encourage future research to further explore and improve the application of open-weight models in the gaming domain, while also leveraging advancements in these models to enhance the capabilities of larger, more general-purpose models.

\subsection{Exploring the Complex Interaction Dynamics between Users, GPT, and its Generated Content}

Although new research in 2024 has reported interesting interactions between users and GPT or GPT-generated content for games, there is still a lack of comprehensive studies on user experiences. We believe that user-centered design will play an important role in PCG, MIGDD, and MIG, and we encourage future research on these use cases to provide detailed reports on user experiences.

\textbf{PCG.} We found that algorithmic benchmarks and metrics are still the primary methods for evaluating GPT's output quality in PCG, particularly for non-text content. In the broader AI field, over-reliance on quantitative metrics has been criticized for potentially obscuring genuine emotional responses and interactive experiences~\cite{relymetrics}.
Users may also exhibit biases toward AI-generated content~\cite{31,53}, being either overly critical or excessively lenient compared to human-made content. As a result, evaluations based solely on benchmarks or metrics may not accurately capture players' real experiences.
Moreover, while human evaluations and feedback are not novel requirements for AI systems, LLMs can better interpret them than traditional models. For example, users may provide direct text feedback about the difficulties of the levels or the theme of the stories, which can be directly integrated to adjust the model's outputs in the game. This approach fosters an iterative cycle, where the model continually learns to align more closely with user expectations over time~\cite{humanfeedback1,humanfeedback2}. By incorporating these nuanced forms of human feedback, evaluations can more accurately capture the real experiences of players, thereby informing improvements in both system design and the scope of GPT-generated content --- a core objective of the Experience-Driven Procedural Content Generation (EDPCG) approach~\cite{experiencepcg}.

\textbf{MIGDD.} In the MIGDD use case, new research from 2024 suggests that designers at different skill levels may require varying degrees of  assistance~\citereview{129,66}. This is especially important for game design, where fluid collaboration between designers and generative models like GPT can foster richer creativity and accommodate diverse workflows. Offering tailored, dynamic support also ensures that design tools remain accessible, helping maintain motivation and encouraging broader participation in the game design process.
Future MIGDD tool research could focus on user-centered design, offering dynamically adjustable support. This means that tools should not only be adjusted according to the designer’s current support needs~\cite{dynamicscaffold} but also adapt to changes in their abilities over time~\cite{hcexp3,hcexpmethod}.
Future research can also consider the values and personalized needs of users. This might involve providing designers with tools that align with their design values~\cite{tooltocompanion}, ethical frameworks~\cite{valuealignmentcowriting,valuealignmentllm}, or creative styles, thus enabling a more inclusive and sustainable design experience.

\textbf{MIG.} As for MIG, we observed that this has been applied more frequently in serious games in 2024, emphasizing the potential for user-centered design to enhance these experiences. 
GPT’s interactive narratives and real-time feedback enable fun, dynamic, and personalized experiences that adapt to the needs of each player, allowing a wider range of users to engage deeply with serious topics~\cite{AIGBL}, such as climate change~\cite{74}, gender~\cite{72}, and education~\cite{71}. However, despite these advantages, challenges such as hallucination, bias, and safety concerns remain critical considerations. GPT-generated content may inadvertently produce misleading or biased narratives, which can affect the integrity of educational and serious games. Ensuring accuracy and fairness in generated content is essential, particularly in crafting immersive, context-sensitive narratives that resonate with players while avoiding the reinforcement of harmful stereotypes or misinformation. Additionally, balancing educational goals with engaging, user-centered interactions requires careful design to respect player autonomy and preferences~\cite{triadic}. Addressing these challenges will be key to maximizing MIG’s potential for fostering meaningful learning and supporting user-driven discussions on complex issues.

\section{Limitation}
Our review has several limitations. 
First, given the rapid pace of development, new GPT use cases in gaming are constantly emerging, and this survey only captured cases published as peer-reviewed full papers, excluding work-in-progress and not peer-reviewed efforts. However, we argue that the use case categorization derived from existing applications in our study will hold significant reference value. In fact, the categories did not change after expanding our updated review significantly (from 55 to 177 papers). Second, while the selected databases include major conferences and journals in the gaming field, we recognize that other databases, particularly those focused on design, may contain relevant topics that were not included in our review. However, we argue that due to the inherently technical nature of AI, our review primarily focuses on technical research rather than design research. Third, our focus on academic publications does not include an examination of commercial games that utilize GPT technologies. The gaming industry is often leading technical innovations in AI, thus examining the industry would be valuable and necessary for a comprehensive understanding. 
Fourth, while our study primarily describes previous research approaches, it does not directly assess their scope and quality, limiting identifying gaps and opportunities for future research. Fifth, our review is limited to GPT-based generative AI, as we observed these models were almost exclusively used in the past few years. As a result, we excluded other AI models, such as Claude and Gemini, which have become more popular recently and could offer alternative methods for game content generation. Finally, it would be worthwhile to compare LLM-based generative AI with other generative AI techniques, such as diffusion models and evolutionary algorithms. 

To address these limitations, future research should incorporate a broader range of sources, including industry applications and design-focused studies, to provide a more comprehensive perspective on GPT’s role in gaming. Expanding the review to include alternative generative AI techniques will provide a more comprehensive understanding of their technical capabilities and how they can be applied to games. Furthermore, conducting meta-analyses or systematic evaluations of existing studies could enhance our understanding of the strengths and weaknesses of different approaches, ultimately contributing to a more robust foundation for AI-driven game design research.

\section{Conclusion}
We conducted a scoping review of 177 articles on GPT applications in games, with 122 published in 2024. Through this review, we identified various uses of GPT within game research. The applications include using GPT for procedural game content generation, employing GPT in mixed-initiative game design and gameplay processes, leveraging GPT to autonomously play games, and using GPT for game user research.
Based on our findings, we suggest that future studies focus on leveraging smaller language models, reporting and exploring player experiences, developing adaptive game design tools that leverage multiple GPT models for diverse tasks, and exploring GPT's role in more complex games.
This work aims to lay the foundation for further innovative applications of GPT in gaming, enhancing both game development and player experience through advanced AI techniques.

\bibliographystyle{ieeetr}

\begin{thebibliography}{10}

\bibitem{GPT4}
J.~Achiam, S.~Adler, S.~Agarwal, L.~Ahmad, I.~Akkaya, F.~L. Aleman, D.~Almeida, J.~Altenschmidt, S.~Altman, S.~Anadkat, {\em et~al.}, ``{GPT-4} technical report,'' {\em arXiv preprint arXiv:2303.08774}, 2023.

\bibitem{GPTsurvey}
K.~S. Kalyan, ``A survey of {GPT}-3 family large language models including {ChatGPT} and {GPT-4},'' {\em Natural Language Processing Journal}, p.~100048, 2023.

\bibitem{entityextraction}
H.~Sousa, N.~Guimar{\~a}es, A.~Jorge, and R.~Campos, ``{GPT} struct me: Probing {GPT} models on narrative entity extraction,'' in {\em 2023 IEEE International Conference on Web Intelligence and Intelligent Agent Technology (WI-IAT)}, pp.~383--387, IEEE, 2023.

\bibitem{QA1}
S.~Kumari and T.~Pushphavati, ``Question answering and text generation using {BERT} and {GPT}-2 model,'' in {\em Computational Methods and Data Engineering: Proceedings of ICCMDE 2021}, pp.~93--110, Springer, 2022.

\bibitem{QA2}
R.~Nakano, J.~Hilton, S.~Balaji, J.~Wu, L.~Ouyang, C.~Kim, C.~Hesse, S.~Jain, V.~Kosaraju, W.~Saunders, {\em et~al.}, ``Web{GPT}: Browser-assisted question-answering with human feedback,'' {\em arXiv preprint arXiv:2112.09332}, 2021.

\bibitem{textgeneration}
J.~Li, T.~Tang, W.~X. Zhao, J.-Y. Nie, and J.-R. Wen, ``Pretrained language models for text generation: A survey,'' {\em arXiv preprint arXiv:2201.05273}, 2022.

\bibitem{programming}
N.~M.~S. Surameery and M.~Y. Shakor, ``Use chat {GPT} to solve programming bugs,'' {\em International Journal of Information Technology \& Computer Engineering (IJITC) ISSN: 2455-5290}, vol.~3, no.~01, pp.~17--22, 2023.

\bibitem{aiasactive}
D.~Yang, Y.~Zhou, Z.~Zhang, T.~J.-J. Li, and R.~LC, ``{AI} as an active writer: Interaction strategies with generated text in human-ai collaborative fiction writing,'' in {\em Joint Proceedings of the ACM IUI Workshops}, vol.~10, CEUR-WS Team, 2022.

\bibitem{creativitysupport}
T.~Chakrabarty, V.~Padmakumar, F.~Brahman, and S.~Muresan, ``Creativity support in the age of large language models: An empirical study involving emerging writers,'' {\em arXiv preprint arXiv:2309.12570}, 2023.

\bibitem{GPT4Game}
D.~Yang, E.~Kleinman, and C.~Harteveld, ``{GPT} for games: A scoping review (2020-2023),'' in {\em 2024 IEEE Conference on Games (CoG)}, pp.~1--8, 2024.

\bibitem{GPT3}
T.~Brown, B.~Mann, N.~Ryder, M.~Subbiah, J.~D. Kaplan, P.~Dhariwal, A.~Neelakantan, P.~Shyam, G.~Sastry, A.~Askell, {\em et~al.}, ``Language models are few-shot learners,'' {\em Advances in neural information processing systems}, vol.~33, pp.~1877--1901, 2020.

\bibitem{review1}
R.~Gallotta, G.~Todd, M.~Zammit, S.~Earle, A.~Liapis, J.~Togelius, and G.~N. Yannakakis, ``Large language models and games: A survey and roadmap,'' {\em arXiv preprint arXiv:2402.18659}, 2024.

\bibitem{review3}
S.~Hu, T.~Huang, F.~Ilhan, S.~Tekin, G.~Liu, R.~Kompella, and L.~Liu, ``A survey on large language model-based game agents,'' {\em arXiv preprint arXiv:2404.02039}, 2024.

\bibitem{review4}
C.~Hu, Y.~Zhao, Z.~Wang, H.~Du, and J.~Liu, ``Games for artificial intelligence research: A review and perspectives,'' {\em IEEE Transactions on Artificial Intelligence}, 2024.

\bibitem{review2}
P.~Sweetser, ``Large language models and video games: A preliminary scoping review,'' in {\em Proceedings of the 6th ACM Conference on Conversational User Interfaces}, pp.~1--8, 2024.

\bibitem{ACM}
S.~Guha, S.~Steinhardt, S.~I. Ahmed, and C.~Lagoze, ``Following bibliometric footprints: the acm digital library and the evolution of computer science,'' in {\em Proceedings of the 13th ACM/IEEE-CS Joint Conference on Digital Libraries}, JCDL '13, (New York, NY, USA), p.~139–142, Association for Computing Machinery, 2013.

\bibitem{IEEE}
L.~Gotsev and E.~Shoikova, ``An analysis of scientific production in big data knowledge domain on google books, youtube and ieee explore® digital library,'' in {\em Proceedings of the 2020 4th International Conference on Cloud and Big Data Computing}, ICCBDC '20, (New York, NY, USA), p.~10–14, Association for Computing Machinery, 2020.

\bibitem{opencoding}
S.~H. Khandkar, ``Open coding,'' {\em University of Calgary}, vol.~23, no.~2009, 2009.

\bibitem{gpt2vsgpt3}
M.~Zhang and J.~Li, ``A commentary of {GPT}-3 in {MIT} technology review 2021,'' {\em Fundamental Research}, vol.~1, no.~6, pp.~831--833, 2021.

\bibitem{rag}
Y.~Gao, Y.~Xiong, X.~Gao, K.~Jia, J.~Pan, Y.~Bi, Y.~Dai, J.~Sun, and H.~Wang, ``Retrieval-augmented generation for large language models: A survey,'' {\em arXiv preprint arXiv:2312.10997}, 2023.

\bibitem{pcgsurvey}
M.~Hendrikx, S.~Meijer, J.~Van Der~Velden, and A.~Iosup, ``Procedural content generation for games: A survey,'' {\em ACM Transactions on Multimedia Computing, Communications, and Applications (TOMM)}, vol.~9, no.~1, pp.~1--22, 2013.

\bibitem{langchain}
H.~Chase, ``Langchain,'' 2022.
\newblock If you use this software, please cite it as below.

\bibitem{bertscore}
T.~Zhang, V.~Kishore, F.~Wu, K.~Q. Weinberger, and Y.~Artzi, ``Bertscore: Evaluating text generation with bert,'' {\em arXiv preprint arXiv:1904.09675}, 2019.

\bibitem{treeofthoughts}
S.~Yao, D.~Yu, J.~Zhao, I.~Shafran, T.~L. Griffiths, Y.~Cao, and K.~Narasimhan, ``Tree of thoughts: Deliberate problem solving with large language models, 2023,'' {\em URL https://arxiv. org/pdf/2305.10601. pdf}, 2023.

\bibitem{quest}
K.~Y. Kristen, N.~R. Sturtevant, and M.~Guzdial, ``What is a quest?,'' in {\em AIIDE Workshops}, 2020.

\bibitem{angrybirds}
R.~Entertainment, ``Angry birds,'' 2009.
\newblock Video game.

\bibitem{minecraft}
M.~Persson, ``Minecraft,'' 2011.
\newblock Video game.

\bibitem{facs}
P.~Ekman and W.~V. Friesen, ``Facial action coding system,'' {\em Environmental Psychology \& Nonverbal Behavior}, 1978.

\bibitem{LMA}
U.~Bernardet, S.~Fdili~Alaoui, K.~Studd, K.~Bradley, P.~Pasquier, and T.~Schiphorst, ``Assessing the reliability of the laban movement analysis system,'' {\em PloS one}, vol.~14, no.~6, p.~e0218179, 2019.

\bibitem{SBBS}
D.~L. Poole and A.~K. Mackworth, {\em Artificial Intelligence: foundations of computational agents}.
\newblock Cambridge University Press, 2010.

\bibitem{mixedinitiativeinterface}
S.~Deterding, J.~Hook, R.~Fiebrink, M.~Gillies, J.~Gow, M.~Akten, G.~Smith, A.~Liapis, and K.~Compton, ``Mixed-initiative creative interfaces,'' in {\em Proceedings of the 2017 CHI Conference Extended Abstracts on Human Factors in Computing Systems}, pp.~628--635, 2017.

\bibitem{dynamicscaffold}
P.~S. Dhillon, S.~Molaei, J.~Li, M.~Golub, S.~Zheng, and L.~P. Robert, ``Shaping human-{AI} collaboration: Varied scaffolding levels in co-writing with language models,'' in {\em Proceedings of the CHI Conference on Human Factors in Computing Systems}, CHI '24, (New York, NY, USA), Association for Computing Machinery, 2024.

\bibitem{vgdl}
T.~Schaul, ``A video game description language for model-based or interactive learning,'' in {\em 2013 IEEE Conference on Computational Inteligence in Games (CIG)}, pp.~1--8, IEEE, 2013.

\bibitem{drl}
H.~Dong, H.~Dong, Z.~Ding, S.~Zhang, and T.~Chang, {\em Deep Reinforcement Learning}.
\newblock Springer, 2020.

\bibitem{lcb}
S.~Toshniwal, S.~Wiseman, K.~Livescu, and K.~Gimpel, ``Chess as a testbed for language model state tracking,'' in {\em Proceedings of the AAAI Conference on Artificial Intelligence}, vol.~36, pp.~11385--11393, 2022.

\bibitem{semeval}
S.~Negi, T.~Daudert, and P.~Buitelaar, ``Semeval-2019 task 9: Suggestion mining from online reviews and forums,'' in {\em Proceedings of the 13th international workshop on semantic evaluation}, pp.~877--887, 2019.

\bibitem{pxi}
V.~V. Abeele, K.~Spiel, L.~Nacke, D.~Johnson, and K.~Gerling, ``Development and validation of the player experience inventory: A scale to measure player experiences at the level of functional and psychosocial consequences,'' {\em International Journal of Human-Computer Studies}, vol.~135, p.~102370, 2020.

\bibitem{rts}
S.~Huang, S.~Onta{\~n}{\'o}n, C.~Bamford, and L.~Grela, ``Gym-$\mu$rts: Toward affordable full game real-time strategy games research with deep reinforcement learning,'' in {\em 2021 IEEE Conference on Games (CoG)}, pp.~1--8, IEEE, 2021.

\bibitem{starcraft2}
{Blizzard Entertainment}, ``Starcraft ii: Wings of liberty,'' 2010.

\bibitem{ageofempires}
{Ensemble Studios}, ``Age of empires,'' 1997.

\bibitem{dontstarvetogether}
{Klei Entertainment}, ``Don't starve together,'' 2016.

\bibitem{ffxiv2010}
{Square Enix}, ``Final fantasy xiv,'' 2010.

\bibitem{overwatch2016}
{Blizzard Entertainment}, ``Overwatch,'' 2016.

\bibitem{CoT}
J.~Wei, X.~Wang, D.~Schuurmans, M.~Bosma, b.~ichter, F.~Xia, E.~Chi, Q.~V. Le, and D.~Zhou, ``Chain-of-thought prompting elicits reasoning in large language models,'' in {\em Advances in Neural Information Processing Systems} (S.~Koyejo, S.~Mohamed, A.~Agarwal, D.~Belgrave, K.~Cho, and A.~Oh, eds.), vol.~35, pp.~24824--24837, Curran Associates, Inc., 2022.

\bibitem{gamedesignmulti}
G.~Smith, R.~Anderson, B.~Kopleck, Z.~Lindblad, L.~Scott, A.~Wardell, J.~Whitehead, and M.~Mateas, ``Situating quests: Design patterns for quest and level design in role-playing games,'' in {\em Interactive Storytelling: Fourth International Conference on Interactive Digital Storytelling, ICIDS 2011, Vancouver, Canada, November 28--1 December, 2011. Proceedings 4}, pp.~326--329, Springer, 2011.

\bibitem{Brainstorming1}
L.-C. Lu, S.-J. Chen, T.-M. Pai, C.-H. Yu, H.-y. Lee, and S.-H. Sun, ``Llm discussion: Enhancing the creativity of large language models via discussion framework and role-play,'' {\em arXiv preprint arXiv:2405.06373}, 2024.

\bibitem{Brainstorming2}
Y.~Liu, P.~Sharma, M.~J. Oswal, H.~Xia, and Y.~Huang, ``Personaflow: Boosting research ideation with llm-simulated expert personas,'' {\em arXiv preprint arXiv:2409.12538}, 2024.

\bibitem{chatdev}
C.~Qian, X.~Cong, C.~Yang, W.~Chen, Y.~Su, J.~Xu, Z.~Liu, and M.~Sun, ``Communicative agents for software development,'' {\em arXiv preprint arXiv:2307.07924}, 2023.

\bibitem{gamegpt}
D.~Chen, H.~Wang, Y.~Huo, Y.~Li, and H.~Zhang, ``Game{GPT}: Multi-agent collaborative framework for game development,'' {\em arXiv preprint arXiv:2310.08067}, 2023.

\bibitem{smallsafe}
K.~Zhang, J.~Wang, E.~Hua, B.~Qi, N.~Ding, and B.~Zhou, ``Cogenesis: A framework collaborating large and small language models for secure context-aware instruction following,'' {\em arXiv preprint arXiv:2403.03129}, 2024.

\bibitem{smallkd}
Y.~Tian, Y.~Han, X.~Chen, W.~Wang, and N.~V. Chawla, ``Tinyllm: Learning a small student from multiple large language models,'' {\em arXiv preprint arXiv:2402.04616}, 2024.

\bibitem{smalltrim}
S.~Muralidharan, S.~T. Sreenivas, R.~Joshi, M.~Chochowski, M.~Patwary, M.~Shoeybi, B.~Catanzaro, J.~Kautz, and P.~Molchanov, ``Compact language models via pruning and knowledge distillation,'' {\em arXiv preprint arXiv:2407.14679}, 2024.

\bibitem{smallplugin}
C.~Xu, Y.~Xu, S.~Wang, Y.~Liu, C.~Zhu, and J.~McAuley, ``Small models are valuable plug-ins for large language models,'' {\em arXiv preprint arXiv:2305.08848}, 2023.

\bibitem{relymetrics}
R.~L. Thomas and D.~Uminsky, ``Reliance on metrics is a fundamental challenge for {AI},'' {\em Patterns}, vol.~3, no.~5, 2022.

\bibitem{humanfeedback1}
C.~Malaviya, S.~Lee, D.~Roth, and M.~Yatskar, ``What if you said that differently?: How explanation formats affect human feedback efficacy and user perception,'' {\em arXiv preprint arXiv:2311.09558}, 2023.

\bibitem{humanfeedback2}
J.~Scheurer, J.~A. Campos, T.~Korbak, J.~S. Chan, A.~Chen, K.~Cho, and E.~Perez, ``Training language models with language feedback at scale,'' {\em arXiv preprint arXiv:2303.16755}, 2023.

\bibitem{experiencepcg}
G.~N. Yannakakis and J.~Togelius, ``Experience-driven procedural content generation,'' {\em IEEE Transactions on Affective Computing}, vol.~2, no.~3, pp.~147--161, 2011.

\bibitem{hcexp3}
Y.~Rong, T.~Leemann, T.-T. Nguyen, L.~Fiedler, P.~Qian, V.~Unhelkar, T.~Seidel, G.~Kasneci, and E.~Kasneci, ``Towards human-centered explainable {AI}: A survey of user studies for model explanations,'' {\em IEEE Transactions on Pattern Analysis and Machine Intelligence}, vol.~46, no.~4, pp.~2104--2122, 2024.

\bibitem{hcexpmethod}
Q.~V. Liao and K.~R. Varshney, ``Human-centered explainable {AI} ({XAI}): From algorithms to user experiences,'' {\em arXiv preprint arXiv:2110.10790}, 2021.

\bibitem{tooltocompanion}
O.~C. Biermann, N.~F. Ma, and D.~Yoon, ``From tool to companion: Storywriters want {AI} writers to respect their personal values and writing strategies,'' in {\em Proceedings of the 2022 ACM Designing Interactive Systems Conference}, DIS '22, (New York, NY, USA), p.~1209–1227, Association for Computing Machinery, 2022.

\bibitem{valuealignmentcowriting}
A.~Guo, P.~Pataranutaporn, and P.~Maes, ``Exploring the impact of {AI} value alignment in collaborative ideation: Effects on perception, ownership, and output,'' in {\em Extended Abstracts of the 2024 CHI Conference on Human Factors in Computing Systems}, CHI EA '24, (New York, NY, USA), Association for Computing Machinery, 2024.

\bibitem{valuealignmentllm}
X.~Yi, J.~Yao, X.~Wang, and X.~Xie, ``Unpacking the ethical value alignment in big models,'' {\em arXiv preprint arXiv:2310.17551}, 2023.

\bibitem{AIGBL}
Y.~Y. Dyulicheva and A.~O. Glazieva, ``Game based learning with artificial intelligence and immersive technologies: An overview.,'' {\em CS\&SE@ SW}, pp.~146--159, 2021.

\bibitem{triadic}
C.~Harteveld, {\em Triadic Game Design: Balancing Reality, Meaning and Play}.
\newblock Springer Science \& Business Media, 2011.


\section*{GPT Game Literature}

\bibitem{14}
F.~Gao, K.~Fang, and W.~K.~V. Chan, ``Chemical life: Knowledge-based personality, emotion and action cues in educational games,'' in {\em 2023 IEEE Conference on Games (CoG)}, pp.~1--3, IEEE, 2023.

\bibitem{7}
P.~Ammanabrolu, W.~Cheung, D.~Tu, W.~Broniec, and M.~Riedl, ``Bringing stories alive: Generating interactive fiction worlds,'' in {\em Proceedings of the AAAI Conference on Artificial Intelligence and Interactive Digital Entertainment}, vol.~16, pp.~3--9, 2020.

\bibitem{37}
M.~G. Torii, T.~Murakami, and Y.~Ochiai, ``Lottery and sprint: Generate a board game with design sprint method on {A}uto{GPT},'' in {\em Companion Proceedings of the Annual Symposium on Computer-Human Interaction in Play}, pp.~259--265, 2023.

\bibitem{93}
A.~Alkhayat, B.~Ciranni, R.~S. Tumuluri, and R.~S. Tulasi, ``Leveraging large language models for enhanced vr development: Insights and challenges,'' in {\em 2024 IEEE Gaming, Entertainment, and Media Conference (GEM)}, pp.~1--6, 2024.

\bibitem{102}
P.~Babiuch, A.~Łapczyński, H.~Jegierski, M.~Jegierski, R.~Salamon, B.~Kolber-Bugajska, M.~Płaza, S.~Deniziak, P.~Pięta, G.~Łukawski, A.~Jasiński, J.~Opałka, A.~Marmon, K.~Kwiatkowski, A.~Cybulski, M.~Igras-Cybulska, and P.~Węgrzyn, ``{MUN}: an {AI}-powered multiplayer networking solution for {VR} games,'' in {\em 2024 IEEE Conference on Virtual Reality and 3D User Interfaces Abstracts and Workshops (VRW)}, pp.~1204--1205, 2024.

\bibitem{35}
Y.~Sun, Z.~Li, K.~Fang, C.~H. Lee, and A.~Asadipour, ``Language as reality: a co-creative storytelling game experience in 1001 nights using generative {AI},'' in {\em Proceedings of the AAAI Conference on Artificial Intelligence and Interactive Digital Entertainment}, vol.~19, pp.~425--434, 2023.

\bibitem{10}
A.~Zhu, L.~Martin, A.~Head, and C.~Callison-Burch, ``{CALYPSO}: {LLM}s as dungeon master's assistants,'' in {\em Proceedings of the AAAI Conference on Artificial Intelligence and Interactive Digital Entertainment}, vol.~19, pp.~380--390, 2023.

\bibitem{122}
F.~Abdullah, P.~Taveekitworachai, M.~F. Dewantoro, R.~Thawonmas, J.~Togelius, and J.~Renz, ``The 1st {C}hat{GPT4PCG} competition,'' {\em IEEE Transactions on Games}, pp.~1--17, 2024.

\bibitem{147}
B.~Yoo and K.-J. Kim, ``Finding deceivers in social context with large language models and how to find them: the case of the mafia game,'' {\em Scientific Reports}, vol.~14, no.~1, p.~30946, 2024.

\bibitem{149}
D.~Li, S.~S. Sohn, S.~Zhang, C.-J. Chang, and M.~Kapadia, ``From words to worlds: Transforming one-line prompts into multi-modal digital stories with {LLM} agents,'' in {\em Proceedings of the 17th ACM SIGGRAPH Conference on Motion, Interaction, and Games}, MIG '24, (New York, NY, USA), Association for Computing Machinery, 2024.

\bibitem{98}
F.~De~La~Torre, C.~M. Fang, H.~Huang, A.~Banburski-Fahey, J.~Amores~Fernandez, and J.~Lanier, ``{LLMR}: Real-time prompting of interactive worlds using large language models,'' in {\em Proceedings of the 2024 CHI Conference on Human Factors in Computing Systems}, CHI '24, (New York, NY, USA), Association for Computing Machinery, 2024.

\bibitem{34}
P.~Taveekitworachai, M.~C. Gursesli, F.~Abdullah, S.~Chen, F.~Cala, A.~Guazzini, A.~Lanata, and R.~Thawonmas, ``Journey of {ChatGPT} from prompts to stories in games: the positive, the negative, and the neutral,'' in {\em 2023 IEEE 13th International Conference on Consumer Electronics-Berlin (ICCE-Berlin)}, pp.~202--203, IEEE, 2023.

\bibitem{38}
Q.~Kuang, F.~Shen, C.~M. Fang, and A.~Dong, ``Memeopoly: An {AI}-powered physical board game interface for tangible play and learning art and design,'' in {\em Companion Proceedings of the Annual Symposium on Computer-Human Interaction in Play}, pp.~292--297, 2023.

\bibitem{15}
R.~LC and V.~Ruijters, ``Chikyuchi: In-person/remote game exhibition for climate change influence,'' in {\em Sixteenth International Conference on Tangible, Embedded, and Embodied Interaction}, pp.~1--4, 2022.

\bibitem{46}
B.~Thabet, N.~Zanichelli, and F.~Zanichelli, ``Q{\&}{A} generation for flashcards within a transformer-based framework,'' in {\em Higher Education Learning Methodologies and Technologies Online} (G.~Fulantelli, D.~Burgos, G.~Casalino, M.~Cimitile, G.~Lo~Bosco, and D.~Taibi, eds.), (Cham), pp.~789--806, Springer Nature Switzerland, 2023.

\bibitem{23}
Y.~Mori and Y.~Miyake, ``Ethical issues in automatic dialogue generation for non-player characters in digital games,'' in {\em 2022 IEEE International Conference on Big Data (Big Data)}, pp.~5132--5139, IEEE, 2022.

\bibitem{28}
Q.~R. Yong and A.~Mitchell, ``From playing the story to gaming the system: Repeat experiences of a large language model-based interactive story,'' in {\em International Conference on Interactive Digital Storytelling}, pp.~395--409, Springer, 2023.

\bibitem{22}
W.~Hettmann, M.~W{\"o}lfel, M.~Butz, K.~Torner, and J.~Finken, ``Engaging museum visitors with {AI}-generated narration and gameplay,'' in {\em International Conference on ArtsIT, Interactivity and Game Creation}, pp.~201--214, Springer, 2022.

\bibitem{54}
P.~Taveekitworachai, F.~Abdullah, M.~C. Gursesli, M.~F. Dewantoro, S.~Chen, A.~Lanata, A.~Guazzini, and R.~Thawonmas, ``What is waiting for us at the end? inherent biases of game story endings in large language models,'' in {\em Interactive Storytelling} (L.~Holloway-Attaway and J.~T. Murray, eds.), (Cham), pp.~274--284, Springer Nature Switzerland, 2023.

\bibitem{20}
K.~Saito, K.~Kobayashi, W.~Takekoshi, A.~Hashimoto, N.~Hirai, A.~Kimura, A.~Takahashi, N.~Yoshioka, and A.~Mano, ``Double impact: Children’s serious {RPG} generation/play with a large language model for their deeper engagement in social issues,'' in {\em Joint International Conference on Serious Games}, pp.~274--289, Springer, 2023.

\bibitem{25}
T.~Triyason, ``Exploring the potential of {ChatGPT} as a dungeon master in dungeons \& dragons tabletop game,'' in {\em Proceedings of the 13th International Conference on Advances in Information Technology}, pp.~1--6, 2023.

\bibitem{26}
C.~Ang, L.~R. Cortel, C.~L. Santos, and E.~Ong, ``Fable reborn: Investigating gameplay experience between a human player and a virtual dungeon master,'' in {\em Extended Abstracts of the 2023 CHI Conference on Human Factors in Computing Systems}, pp.~1--7, 2023.

\bibitem{57}
H.~Tang and M.~Singha, ``A mystery for you: A fact-checking game enhanced by large language models ({LLMs}) and a tangible interface,'' in {\em Extended Abstracts of the 2024 CHI Conference on Human Factors in Computing Systems}, CHI EA '24, (New York, NY, USA), Association for Computing Machinery, 2024.

\bibitem{58}
P.~Mieschke and S.~Radicke, ``Adaptive {LLM}-based game radio ({ALGR}),'' in {\em 2023 4th International Conference on Computers and Artificial Intelligence Technology (CAIT)}, pp.~277--283, 2023.

\bibitem{61}
P.~Taveekitworachai, K.~Plupattanakit, and R.~Thawonmas, ``Assessing inherent biases following prompt compression of large language models for game story generation,'' in {\em 2024 IEEE Conference on Games (CoG)}, pp.~1--4, 2024.

\bibitem{63}
W.~Du, Z.~Zhu, X.~Xu, H.~Che, and S.~Chen, ``Careersim: Gamification design leveraging {LLM}s for career development reflection,'' in {\em Extended Abstracts of the 2024 CHI Conference on Human Factors in Computing Systems}, CHI EA '24, (New York, NY, USA), Association for Computing Machinery, 2024.

\bibitem{69}
S.~R. Cox and W.~T. Ooi, ``Conversational interactions with npcs in {LLM}-driven gaming: Guidelines from a content analysis of player feedback,'' in {\em International Workshop on Chatbot Research and Design}, pp.~167--184, Springer, 2023.

\bibitem{86}
P.~Taveekitworachai, M.~C. Gursesli, F.~Abdullah, S.~Chen, F.~Cala, A.~Guazzini, A.~Lanata, and R.~Thawonmas, ``Journey of {C}hat{GPT} from prompts to stories in games: the positive, the negative, and the neutral,'' in {\em 2023 IEEE 13th International Conference on Consumer Electronics - Berlin (ICCE-Berlin)}, pp.~202--203, 2023.

\bibitem{92}
S.~Buongiorno and C.~Clark, ``Leveraging gaming to enhance knowledge graphs for explainable generative {AI} applications,'' in {\em 2024 IEEE Conference on Games (CoG)}, pp.~1--4, 2024.

\bibitem{123}
A.~Normoyle, S.~J\"{o}rg, and J.~Hill, ``The curation tree: A lightweight behavior tree framework for implementing puzzle and narrative games,'' in {\em Proceedings of the 19th International Conference on the Foundations of Digital Games}, FDG '24, (New York, NY, USA), Association for Computing Machinery, 2024.

\bibitem{91}
T.~Yu, M.~Chen, Y.~Li, D.~Lew, and K.~Yu, ``La{S}ofa: Integrating fantasy storytelling in human-robot interaction through an interactive sofa robot,'' in {\em Companion of the 2024 ACM/IEEE International Conference on Human-Robot Interaction}, HRI '24, (New York, NY, USA), p.~1168–1172, Association for Computing Machinery, 2024.

\bibitem{159}
R.~Farrell and S.~G. Ware, ``Large language models as narrative planning search guides,'' {\em IEEE Transactions on Games}, pp.~1--10, 2024.

\bibitem{135}
R.~Y. Camilleri and V.~Camilleri, ``Beyond the maze: How {AI} personalizes learning and drives engagement in educational games,'' in {\em Proceedings of the Future Technologies Conference (FTC) 2024, Volume 2} (K.~Arai, ed.), (Cham), pp.~301--320, Springer Nature Switzerland, 2024.

\bibitem{163}
T.~Merino, S.~Earle, R.~Sudhakaran, S.~Sudhakaran, and J.~Togelius, ``Making new connections: {LLM}s as puzzle generators for the new york times’ connections word game,'' {\em Proceedings of the AAAI Conference on Artificial Intelligence and Interactive Digital Entertainment}, vol.~20, pp.~87--96, Nov. 2024.

\bibitem{27}
J.~van Stegeren and J.~My{\'s}liwiec, ``Fine-tuning {GPT}-2 on annotated {RPG} quests for {NPC} dialogue generation,'' in {\em Proceedings of the 16th International Conference on the Foundations of Digital Games}, pp.~1--8, 2021.

\bibitem{41}
T.~Ashby, B.~K. Webb, G.~Knapp, J.~Searle, and N.~Fulda, ``Personalized quest and dialogue generation in role-playing games: A knowledge graph- and language model-based approach,'' in {\em Proceedings of the 2023 CHI Conference on Human Factors in Computing Systems}, CHI '23, (New York, NY, USA), Association for Computing Machinery, 2023.

\bibitem{47}
S.~Al-Nassar, A.~Schaap, M.~V.~D. Zwart, M.~Preuss, and M.~A. G\'{o}mez-Maureira, ``Questville: Procedural quest generation using {NLP} models,'' in {\em Proceedings of the 18th International Conference on the Foundations of Digital Games}, FDG '23, (New York, NY, USA), Association for Computing Machinery, 2023.

\bibitem{31}
S.~V{\"a}rtinen, P.~H{\"a}m{\"a}l{\"a}inen, and C.~Guckelsberger, ``Generating role-playing game quests with {GPT} language models,'' {\em IEEE Transactions on Games}, 2022.

\bibitem{45}
S.~Sudhakaran, M.~Gonz\'{a}lez-Duque, C.~Glanois, M.~Freiberger, E.~Najarro, and S.~Risi, ``Prompt-guided level generation,'' in {\em Proceedings of the Companion Conference on Genetic and Evolutionary Computation}, GECCO '23 Companion, (New York, NY, USA), p.~179–182, Association for Computing Machinery, 2023.

\bibitem{36}
G.~Todd, S.~Earle, M.~U. Nasir, M.~C. Green, and J.~Togelius, ``Level generation through large language models,'' in {\em Proceedings of the 18th International Conference on the Foundations of Digital Games}, FDG '23, (New York, NY, USA), Association for Computing Machinery, 2023.

\bibitem{43}
M.~U. Nasir and J.~Togelius, ``Practical {PCG} through large language models,'' in {\em 2023 IEEE Conference on Games (CoG)}, pp.~1--4, 2023.

\bibitem{56}
S.~Hu, Z.~Huang, C.~Hu, and J.~Liu, ``3d building generation in minecraft via large language models,'' in {\em 2024 IEEE Conference on Games (CoG)}, pp.~1--4, 2024.

\bibitem{132}
D.~Hafnar and J.~Demšar, ``Zero-shot reasoning: Personalized content generation without the cold start problem,'' {\em IEEE Transactions on Games}, pp.~1--10, 2024.

\bibitem{1}
T.-H. Fu and K.-C. Wu, ``A dynamic fitness game content generation system based on machine learning,'' in {\em Artificial Intelligence in HCI} (H.~Degen and S.~Ntoa, eds.), (Cham), pp.~50--62, Springer Nature Switzerland, 2023.

\bibitem{21}
J.-Y. Zhou and T.-H. Fu, ``Empowering interactive fitness game,'' in {\em 2023 International Conference on Consumer Electronics - Taiwan (ICCE-Taiwan)}, pp.~279--280, 2023.

\bibitem{101}
S.~Zheng, K.~He, L.~Yang, and J.~Xiong, ``Memory{R}epository for {AI} {NPC},'' {\em IEEE Access}, vol.~12, pp.~62581--62596, 2024.

\bibitem{130}
A.~Normoyle, J.~Sedoc, and F.~Durupinar, ``Using {LLM}s to animate interactive story characters with emotions and personality,'' in {\em 2024 IEEE Conference on Virtual Reality and 3D User Interfaces Abstracts and Workshops (VRW)}, pp.~632--635, 2024.

\bibitem{106}
X.~Peng, J.~Quaye, S.~Rao, W.~Xu, P.~Botchway, C.~Brockett, N.~Jojic, G.~DesGarennes, K.~Lobb, M.~Xu, J.~Leandro, C.~Jin, and B.~Dolan, ``Player-driven emergence in {LLM}-driven game narrative,'' in {\em 2024 IEEE Conference on Games (CoG)}, pp.~1--8, 2024.

\bibitem{73}
J.~Sissler, ``Enhancing non-player characters in {U}nity 3{D} using {GPT}-3.5,'' {\em ACM Games}, vol.~2, Aug. 2024.

\bibitem{88}
S.~Karaosmanoglu, E.~L. Fittschen, H.~Eyicalis, D.~Kraus, H.~Nickelmann, A.~Tomko, and F.~Steinicke, ``Language of {Z}elda: Facilitating language learning practices using {C}hat{GPT},'' in {\em Extended Abstracts of the 2024 CHI Conference on Human Factors in Computing Systems}, CHI EA '24, (New York, NY, USA), Association for Computing Machinery, 2024.

\bibitem{89}
Y.~Zhao, J.~Pan, Y.~Dong, T.~Dong, G.~Wang, F.~Ying, Q.~Shen, and J.~Cao, ``Language urban odyssey: A serious game for enhancing second language acquisition through large language models,'' in {\em Extended Abstracts of the 2024 CHI Conference on Human Factors in Computing Systems}, CHI EA '24, (New York, NY, USA), Association for Computing Machinery, 2024.

\bibitem{116}
A.~Y. Cheng, M.~Guo, M.~Ran, A.~Ranasaria, A.~Sharma, A.~Xie, K.~N. Le, B.~Vinaithirthan, S.~T. Luan, D.~T.~H. Wright, A.~Cuadra, R.~Pea, and J.~A. Landay, ``Scientific and fantastical: Creating immersive, culturally relevant learning experiences with augmented reality and large language models,'' in {\em Proceedings of the 2024 CHI Conference on Human Factors in Computing Systems}, CHI '24, (New York, NY, USA), Association for Computing Machinery, 2024.

\bibitem{127}
B.~Ngaw, G.~Jena, J.~a. Sedoc, and A.~Normoyle, ``Towards authoring open-ended behaviors for narrative puzzle games with large language model support,'' in {\em Proceedings of the 19th International Conference on the Foundations of Digital Games}, FDG '24, (New York, NY, USA), Association for Computing Machinery, 2024.

\bibitem{99}
M.~Dai, C.~Yuan, and X.~Nie, ``Managing the personality of {NPC}s with your interactions: A game design system based on large language models,'' in {\em International Conference on Human-Computer Interaction}, pp.~247--259, Springer, 2024.

\bibitem{105}
J.~Kelly, M.~Mateas, and N.~Wardrip-Fruin, ``Paradise: An experiment extending the ensemble social physics engine with language models,'' in {\em Proceedings of the 19th International Conference on the Foundations of Digital Games}, FDG '24, (New York, NY, USA), Association for Computing Machinery, 2024.

\bibitem{121}
Q.~Sun, Q.~Luo, Y.~Ni, and H.~Mi, ``Text2{AC}: A framework for game-ready 2{D} agent character({AC}) generation from natural language,'' in {\em Extended Abstracts of the 2024 CHI Conference on Human Factors in Computing Systems}, CHI EA '24, (New York, NY, USA), Association for Computing Machinery, 2024.

\bibitem{104}
R.~Zhao, W.~Zhang, J.~Li, L.~Zhu, Y.~Li, Y.~He, and L.~Gui, ``Narrative{P}lay: An automated system for crafting visual worlds in novels for role-playing,'' {\em Proceedings of the AAAI Conference on Artificial Intelligence}, vol.~38, pp.~23859--23861, Mar. 2024.

\bibitem{133}
M.~C. Uludagli and K.~Oguz, ``A social network generator for games evaluated against a real {NPC} network with {GPT}-generated node attributes,'' in {\em 2024 Innovations in Intelligent Systems and Applications Conference (ASYU)}, pp.~1--5, 2024.

\bibitem{145}
L.~J. Klinkert, S.~Buongiorno, and C.~Clark, ``Evaluating the efficacy of {LLM}s to emulate realistic human personalities,'' {\em Proceedings of the AAAI Conference on Artificial Intelligence and Interactive Digital Entertainment}, vol.~20, pp.~65--75, Nov. 2024.

\bibitem{165}
C.~Gr{\'e}visse, ``Ras{P}atient pi: A low-cost customizable {LLM}-based virtual standardized patient simulator,'' in {\em Applied Informatics} (H.~Florez and H.~Astudillo, eds.), (Cham), pp.~125--137, Springer Nature Switzerland, 2025.

\bibitem{142}
K.~Tsuyuguchi, K.~Shimizu, and K.~Suzuki, ``Emotion overflow: an interactive system to represent emotion with fluid,'' in {\em Adjunct Proceedings of the 37th Annual ACM Symposium on User Interface Software and Technology}, UIST Adjunct '24, (New York, NY, USA), Association for Computing Machinery, 2024.

\bibitem{167}
S.~S. Maram, Y.~Malegaonkar, M.~Escarce~Junior, and M.~Seif El-Nasr, ``Shloka: Developing climate change interventions through a lens of religion and videogames,'' in {\em Companion Proceedings of the 2024 Annual Symposium on Computer-Human Interaction in Play}, CHI PLAY Companion '24, (New York, NY, USA), p.~181–187, Association for Computing Machinery, 2024.

\bibitem{17}
L.~Ferreira, L.~Lelis, and J.~Whitehead, ``Computer-generated music for tabletop role-playing games,'' in {\em Proceedings of the AAAI Conference on Artificial Intelligence and Interactive Digital Entertainment}, vol.~16, pp.~59--65, 2020.

\bibitem{5}
C.~Nimpattanavong, P.~Taveekitworachai, I.~Khan, T.~V. Nguyen, R.~Thawonmas, W.~Choensawat, and K.~Sookhanaphibarn, ``Am {I} fighting well? fighting game commentary generation with {ChatGPT},'' in {\em Proceedings of the 13th International Conference on Advances in Information Technology}, pp.~1--7, 2023.

\bibitem{39}
R.~AlJammaz, M.~Mateas, and N.~Wardrip-Fruin, ``Modeling morality-based argumentation for believable game characters: a design postmortem,'' in {\em Proceedings of the AAAI Conference on Artificial Intelligence and Interactive Digital Entertainment}, vol.~19, pp.~185--194, 2023.

\bibitem{30}
F.~Horn, S.~Vogt, and S.~P. G{\"o}bel, ``Game{TUL}earn: An interactive educational game authoring tool for 3{D} environments,'' in {\em Joint International Conference on Serious Games}, pp.~384--390, Springer, 2023.

\bibitem{49}
V.~Kumaran, J.~Rowe, B.~Mott, and J.~Lester, ``Scene{C}raft: Automating interactive narrative scene generation in digital games with large language models,'' {\em Proceedings of the AAAI Conference on Artificial Intelligence and Interactive Digital Entertainment}, vol.~19, pp.~86--96, Oct. 2023.

\bibitem{50}
L.~R. Kouzelis and O.~Spantidi, ``Synthesizing play-ready {VR} scenes with natural language prompts through {GPT} {API},'' in {\em Advances in Visual Computing} (G.~Bebis, G.~Ghiasi, Y.~Fang, A.~Sharf, Y.~Dong, C.~Weaver, Z.~Leo, J.~J. LaViola~Jr., and L.~Kohli, eds.), (Cham), pp.~15--26, Springer Nature Switzerland, 2023.

\bibitem{40}
V.~Burkus, A.~K{\'a}rp{\'a}ti, and L.~Sz{\'e}csi, ``{NLP}-assisted educational memory game experiment,'' in {\em Methodologies and Intelligent Systems for Technology Enhanced Learning, Workshops - 13th International Conference} (Z.~Kubincov{\'a}, F.~Caruso, T.-e. Kim, M.~Ivanova, L.~Lancia, and M.~A. Pellegrino, eds.), (Cham), pp.~59--69, Springer Nature Switzerland, 2023.

\bibitem{107}
B.~Feng, M.~Su, K.~Zeng, and X.~Li, ``Player-oriented procedural generation: Producing desired game content by natural language,'' in {\em International Conference on Human-Computer Interaction}, pp.~260--274, Springer, 2024.

\bibitem{110}
V.~Kumaran, D.~Carpenter, J.~Rowe, B.~Mott, and J.~Lester, ``Procedural level generation in educational games from natural language instruction,'' {\em IEEE Transactions on Games}, pp.~1--10, 2024.

\bibitem{97}
R.~Gallotta, A.~Liapis, and G.~Yannakakis, ``L{LM}aker: A game level design interface using (only) natural language,'' in {\em 2024 IEEE Conference on Games (CoG)}, pp.~1--2, 2024.

\bibitem{109}
X.~Zhang, F.~Wan, K.~Zhang, and H.~Jiang, ``Possible applications of language models in visual novel,'' in {\em 2023 International Conference on Educational Knowledge and Informatization (EKI)}, pp.~60--63, 2023.

\bibitem{67}
R.~Gallotta, A.~Liapis, and G.~Yannakakis, ``Consistent game content creation via function calling for large language models,'' in {\em 2024 IEEE Conference on Games (CoG)}, pp.~1--4, 2024.

\bibitem{75}
D.~Sezen, A.~Akcali, T.~I. Sezen, S.~Perks, and P.~Stewart, ``Exploring women's role in creative industries through collaborative action research using tabletop role-playing games,'' in {\em EAI International Conference on Technology, Innovation, Entrepreneurship and Education}, pp.~39--53, Springer, 2023.

\bibitem{78}
J.~Leandro, S.~Rao, M.~Xu, W.~Xu, N.~Jojic, C.~Brockett, and B.~Dolan, ``{GENEVA}: {GENE}rating and visualizing branching narratives using {LLM}s,'' in {\em 2024 IEEE Conference on Games (CoG)}, pp.~1--5, 2024.

\bibitem{94}
M.~Yin, E.~Wang, C.~Ng, and R.~Xiao, ``Lies, deceit, and hallucinations: Player perception and expectations regarding trust and deception in games,'' in {\em Proceedings of the 2024 CHI Conference on Human Factors in Computing Systems}, CHI '24, (New York, NY, USA), Association for Computing Machinery, 2024.

\bibitem{103}
J.~Li, Z.~Chen, W.~Lin, L.~Zou, X.~Xie, Y.~Hu, and D.~Li, ``Mystery game script compose based on a large language model,'' in {\em 2024 IEEE World AI IoT Congress (AIIoT)}, pp.~451--455, 2024.

\bibitem{134}
D.-A. Ciungan, N.~Goga, I.-A. Bratosin, and R.-C. Popa, ``An intelligent system for image generation in unity,'' in {\em 2024 IEEE SmartBlock4Africa}, pp.~1--5, 2024.

\bibitem{168}
L.~Ling, X.~Chen, R.~Wen, T.~J.-J. Li, and R.~LC, ``Sketchar: Supporting character design and illustration prototyping using generative {AI},'' {\em Proc. ACM Hum.-Comput. Interact.}, vol.~8, Oct. 2024.

\bibitem{169}
E.~M. Yi~Chan, C.~K. Seow, E.~S. Wee~Tan, M.~Wang, P.~C. Yau, and Q.~Cao, ``Sketchboard: Sketch-guided storyboard generation for game characters in the game industry,'' in {\em 2024 IEEE 22nd International Conference on Industrial Informatics (INDIN)}, pp.~1--8, 2024.

\bibitem{173}
F.~Bonetti, A.~Bucchiarone, and V.~Wanick, ``Using {LLM}s to adapt serious games with educators in the loop,'' in {\em Games and Learning Alliance} (A.~Sch{\"o}nbohm, F.~Bellotti, A.~Bucchiarone, F.~de~Rosa, M.~Ninaus, A.~Wang, V.~Wanick, and P.~Dondio, eds.), (Cham), pp.~68--77, Springer Nature Switzerland, 2025.

\bibitem{136}
Z.~Wu and H.~Sato, ``Build-it-here: Utilizing 2{D} inpainting models for on-site 3{D} object generation,'' in {\em 2024 IEEE 9th International Conference on Computational Intelligence and Applications (ICCIA)}, pp.~104--108, 2024.

\bibitem{176}
L.~Zhang, J.~Pan, J.~Gettig, S.~Oney, and A.~Guo, ``{VRC}opilot: Authoring 3d layouts with generative {AI} models in {VR},'' in {\em Proceedings of the 37th Annual ACM Symposium on User Interface Software and Technology}, UIST '24, (New York, NY, USA), Association for Computing Machinery, 2024.

\bibitem{164}
S.~Buongiorno, L.~Klinkert, Z.~Zhuang, T.~Chawla, and C.~Clark, ``{PANG}e{A}: Procedural artificial narrative using generative {AI} for turn-based, role-playing video games,'' {\em Proceedings of the AAAI Conference on Artificial Intelligence and Interactive Digital Entertainment}, vol.~20, pp.~156--166, Nov. 2024.

\bibitem{2}
M.~Charity, Y.~Bhartia, D.~Zhang, A.~Khalifa, and J.~Togelius, ``A preliminary study on a conceptual game feature generation and recommendation system,'' {\em arXiv preprint arXiv:2308.13538}, 2023.

\bibitem{33}
T.~Fulcini and M.~Torchiano, ``Is {ChatGPT} capable of crafting gamification strategies for software engineering tasks?,'' in {\em Proceedings of the 2nd International Workshop on Gamification in Software Development, Verification, and Validation}, pp.~22--28, 2023.

\bibitem{11}
P.~L. Lanzi and D.~Loiacono, ``{C}hat{GPT} and other large language models as evolutionary engines for online interactive collaborative game design,'' GECCO '23, p.~1383–1390, ACM, 2023.

\bibitem{12}
A.~M. Grow and F.~Khosmood, ``{ChatGPT} gamejam: Unleashing the power of large language models for game jams,'' in {\em Proceedings of the 7th International Conference on Game Jams, Hackathons and Game Creation Events}, pp.~51--54, 2023.

\bibitem{76}
C.~Hu, Y.~Zhao, and J.~Liu, ``Game generation via large language models,'' in {\em 2024 IEEE Conference on Games (CoG)}, pp.~1--4, 2024.

\bibitem{124}
A.~Anjum, Y.~Li, N.~Law, M.~Charity, and J.~Togelius, ``The ink splotch effect: A case study on {C}hat{GPT} as a co-creative game designer,'' in {\em Proceedings of the 19th International Conference on the Foundations of Digital Games}, FDG '24, (New York, NY, USA), Association for Computing Machinery, 2024.

\bibitem{151}
T.~Tanaka and E.~Simo-Serra, ``Grammar-based game description generation using large language models,'' {\em IEEE Transactions on Games}, pp.~1--14, 2024.

\bibitem{51}
T.~Merino, R.~Negri, D.~Rajesh, M.~Charity, and J.~Togelius, ``The five-dollar model: Generating game maps and sprites from sentence embeddings,'' {\em Proceedings of the AAAI Conference on Artificial Intelligence and Interactive Digital Entertainment}, vol.~19, pp.~107--115, Oct. 2023.

\bibitem{79}
M.~R. Taesiri, T.~Feng, C.-P. Bezemer, and A.~Nguyen, ``Glitch{B}ench: Can large multimodal models detect video game glitches?,'' in {\em 2024 IEEE/CVF Conference on Computer Vision and Pattern Recognition (CVPR)}, pp.~22444--22455, 2024.

\bibitem{59}
I.~J. Pérez-Colado, M.~Freire-Morán, A.~Calvo-Morata, V.~M. Pérez-Colado, and B.~Fernández-Manjón, ``Ai asyet another tool in undergraduate student projects: Preliminary results,'' in {\em 2024 IEEE Global Engineering Education Conference (EDUCON)}, pp.~1--7, 2024.

\bibitem{65}
I.-C. Baek, T.-H. Park, J.-H. Noh, C.-M. Bae, and K.-J. Kim, ``Chat{PCG}: Large language model-driven reward design for procedural content generation,'' in {\em 2024 IEEE Conference on Games (CoG)}, pp.~1--4, 2024.

\bibitem{128}
L.~Bingli and D.~V. Vargas, ``Towards immersive computational storytelling: Card-framework for enhanced persona-driven dialogues,'' {\em IEEE Transactions on Games}, pp.~1--13, 2024.

\bibitem{172}
P.~Wightman, ``Twisted games: A first experience of inclusion of {AI} tools in first year programming classes,'' in {\em 2024 IEEE Latin American Conference on Computational Intelligence (LA-CCI)}, pp.~1--6, 2024.

\bibitem{9}
H.~Osone, J.-L. Lu, and Y.~Ochiai, ``Buncho: {AI} supported story co-creation via unsupervised multitask learning to increase writers’ creativity in {J}apanese,'' in {\em Extended Abstracts of the 2021 CHI Conference on Human Factors in Computing Systems}, pp.~1--10, 2021.

\bibitem{44}
J.~Freiknecht and W.~Effelsberg, ``Procedural generation of interactive stories using language models,'' in {\em Proceedings of the 15th International Conference on the Foundations of Digital Games}, FDG '20, (New York, NY, USA), Association for Computing Machinery, 2020.

\bibitem{16}
E.~Nichols, L.~Gao, and R.~Gomez, ``Collaborative storytelling with large-scale neural language models,'' in {\em Proceedings of the 13th ACM SIGGRAPH Conference on Motion, Interaction and Games}, pp.~1--10, 2020.

\bibitem{19}
M.~Elgarf, S.~Zojaji, G.~Skantze, and C.~Peters, ``Creativebot: a creative storyteller robot to stimulate creativity in children,'' in {\em Proceedings of the 2022 International Conference on Multimodal Interaction}, pp.~540--548, 2022.

\bibitem{48}
H.~Shakeri, C.~Neustaedter, and S.~DiPaola, ``{SAGA}: Collaborative storytelling with {GPT}-3,'' in {\em Companion Publication of the 2021 Conference on Computer Supported Cooperative Work and Social Computing}, CSCW '21 Companion, (New York, NY, USA), p.~163–166, Association for Computing Machinery, 2021.

\bibitem{24}
B.~M. McLaren, ``Evaluating {{ChatGPT}}’s decimal skills and feedback generation in a digital learning game,'' in {\em Responsive and Sustainable Educational Futures: 18th European Conference on Technology Enhanced Learning, EC-TEL 2023, Aveiro, Portugal, September 4--8, 2023, Proceedings}, vol.~14200, p.~278, Springer Nature, 2023.

\bibitem{8}
Y.~Sun, X.~Ni, H.~Feng, R.~LC, C.~H. Lee, and A.~Asadipour, ``Bringing stories to life in 1001 nights: A co-creative text adventure game using a story generation model,'' in {\em International Conference on Interactive Digital Storytelling}, pp.~651--672, Springer, 2022.

\bibitem{126}
R.~Divanji, A.~Dangol, E.~J. Lombard, K.~Chen, and J.~D. Rubin, ``Togethertales {RPG}: Prosocial skill development through digitally mediated collaborative role-playing,'' in {\em Proceedings of the 23rd Annual ACM Interaction Design and Children Conference}, IDC '24, (New York, NY, USA), p.~1012–1015, Association for Computing Machinery, 2024.

\bibitem{74}
S.~Zhou, L.~B. Hendra, Q.~Zhang, J.~Holopainen, and R.~LC, ``Eternagram: Probing player attitudes towards climate change using a {C}hat{GPT}-driven text-based adventure,'' in {\em Proceedings of the 2024 CHI Conference on Human Factors in Computing Systems}, CHI '24, (New York, NY, USA), Association for Computing Machinery, 2024.

\bibitem{115}
P.~Almeida, A.~Teixeira, A.~Velhinho, R.~Raposo, T.~Silva, and L.~Pedro, ``Remixing and repurposing cultural heritage archives through a collaborative and {AI}-generated storytelling digital platform,'' in {\em Proceedings of the 2024 ACM International Conference on Interactive Media Experiences Workshops}, IMXw '24, (New York, NY, USA), p.~100–104, Association for Computing Machinery, 2024.

\bibitem{118}
D.~Yang, E.~Kleinman, G.~M. Troiano, E.~Tochilnikova, and C.~Harteveld, ``Snake story: Exploring game mechanics for mixed-initiative co-creative storytelling games,'' in {\em Proceedings of the 19th International Conference on the Foundations of Digital Games}, FDG '24, (New York, NY, USA), Association for Computing Machinery, 2024.

\bibitem{120}
Y.~Wang, Q.~Zhou, and D.~Ledo, ``Story{V}erse: Towards co-authoring dynamic plot with {LLM}-based character simulation via narrative planning,'' in {\em Proceedings of the 19th International Conference on the Foundations of Digital Games}, FDG '24, (New York, NY, USA), Association for Computing Machinery, 2024.

\bibitem{62}
B.~Lyman, A.~Ebrahimi, J.~Cox, S.~Chan, C.~Barney, and B.~De~Schutter, ``Cardistry: Exploring a gpt model workflow as an adapted method of gaminiscing,'' in {\em Proceedings of the 19th International Conference on the Foundations of Digital Games}, FDG '24, (New York, NY, USA), Association for Computing Machinery, 2024.

\bibitem{64}
E.~S. De~Lima, B.~Feij\'{o}, M.~A. Cassanova, and A.~L. Furtado, ``Chat{G}eppetto - an {AI}-powered storyteller,'' in {\em Proceedings of the 22nd Brazilian Symposium on Games and Digital Entertainment}, SBGames '23, (New York, NY, USA), p.~28–37, Association for Computing Machinery, 2024.

\bibitem{66}
C.~Lee and J.~R. Mindel, ``Closer and closer worlds: Using {LLM}s to surface personal stories in world-building conversation games,'' in {\em Companion Publication of the 2024 ACM Designing Interactive Systems Conference}, DIS '24 Companion, (New York, NY, USA), p.~289–293, Association for Computing Machinery, 2024.

\bibitem{72}
T.~Reichert, M.~Miftari, C.~Herling, and N.~Marsden, ``Empowering female founders with {AI} and play: Integration of a large language model into a serious game with player-generated content,'' in {\em International Conference on Human-Computer Interaction}, pp.~69--83, Springer, 2024.

\bibitem{100}
C.~Zhang, X.~Liu, K.~Ziska, S.~Jeon, C.-L. Yu, and Y.~Xu, ``Mathemyths: Leveraging large language models to teach mathematical language through child-{AI} co-creative storytelling,'' in {\em Proceedings of the 2024 CHI Conference on Human Factors in Computing Systems}, CHI '24, (New York, NY, USA), Association for Computing Machinery, 2024.

\bibitem{81}
M.~Valentim, J.~Falk, and N.~Inie, ``Hacc-man: An arcade game for jailbreaking llms,'' in {\em Companion Publication of the 2024 ACM Designing Interactive Systems Conference}, DIS '24 Companion, (New York, NY, USA), p.~338–341, Association for Computing Machinery, 2024.

\bibitem{138}
J.~P. Vieira~Sousa, P.~Campos, and P.~Bala, ``College tales: Pilot study on large language models generated narratives for mental health literacy,'' in {\em Proceedings of the 27th International Academic Mindtrek Conference}, Mindtrek '24, (New York, NY, USA), p.~270–275, Association for Computing Machinery, 2024.

\bibitem{140}
A.~B\"{o}r\"{u}tecene and O.~O. Buruk, ``Dr. ping and dr. pong: Rethinking writing and work with playful embodied ais,'' in {\em Proceedings of the Halfway to the Future Symposium}, HttF '24, (New York, NY, USA), Association for Computing Machinery, 2024.

\bibitem{158}
V.~Shrivastava, S.~Sharma, D.~Chakraborty, and M.~Kinnula, ``Is a sunny day bright and cheerful or hot and uncomfortable? young children's exploration of chat{GPT},'' in {\em Proceedings of the 13th Nordic Conference on Human-Computer Interaction}, NordiCHI '24, (New York, NY, USA), Association for Computing Machinery, 2024.

\bibitem{144}
T.~Tucek, ``Enhancing empathy through personalized {AI}-driven experiences and conversations with digital humans in video games,'' in {\em Companion Proceedings of the 2024 Annual Symposium on Computer-Human Interaction in Play}, CHI PLAY Companion '24, (New York, NY, USA), p.~446–449, Association for Computing Machinery, 2024.

\bibitem{141}
Y.~Lyu, D.~Liu, P.~An, X.~Tong, H.~Zhang, K.~Katsuragawa, and J.~Zhao, ``Emooly: Supporting autistic children in collaborative social-emotional learning with caregiver participation through interactive {AI}-infused and {AR} activities,'' {\em Proc. ACM Interact. Mob. Wearable Ubiquitous Technol.}, vol.~8, Nov. 2024.

\bibitem{146}
F.~R. Christiansen, L.~N. Hollensberg, N.~B. Jensen, K.~Julsgaard, K.~N. Jespersen, and I.~Nikolov, ``Exploring presence in interactions with {LLM}-driven {NPC}s: A comparative study of speech recognition and dialogue options,'' in {\em Proceedings of the 30th ACM Symposium on Virtual Reality Software and Technology}, VRST '24, (New York, NY, USA), Association for Computing Machinery, 2024.

\bibitem{155}
F.~Gao, K.~Fang, and W.~K.~V. Chan, ``Humanizing artifacts: An educational game for cultural heritage artifacts and history using generative {AI},'' in {\em Companion Proceedings of the 2024 Annual Symposium on Computer-Human Interaction in Play}, CHI PLAY Companion '24, (New York, NY, USA), p.~91–96, Association for Computing Machinery, 2024.

\bibitem{177}
N.~Jennings, H.~Wang, I.~Li, J.~Smith, and B.~Hartmann, ``What's the game, then? opportunities and challenges for runtime behavior generation,'' in {\em Proceedings of the 37th Annual ACM Symposium on User Interface Software and Technology}, UIST '24, (New York, NY, USA), Association for Computing Machinery, 2024.

\bibitem{150}
A.~Phillips, J.~Lang, and D.~Mould, ``Goal-oriented interactions in games using {LLM}s,'' {\em IEEE Transactions on Games}, pp.~1--12, 2024.

\bibitem{18}
L.~Huang and X.~Sun, ``Create ice cream: Real-time creative element synthesis framework based on gpt3. 0,'' in {\em 2023 IEEE Conference on Games (CoG)}, pp.~1--4, IEEE, 2023.

\bibitem{114}
A.~Isaza-Giraldo, P.~Bala, P.~F. Campos, and L.~Pereira, ``Prompt-gaming: A pilot study on {LLM}-evaluating agent in a meaningful energy game,'' in {\em Extended Abstracts of the 2024 CHI Conference on Human Factors in Computing Systems}, CHI EA '24, (New York, NY, USA), Association for Computing Machinery, 2024.

\bibitem{111}
M.~Pears, C.~Poussa, and S.~T. Konstantinidis, ``Progressive healthcare pedagogy: An application merging {C}hat{GPT} and {AI}-video technologies for gamified and cost-effective scenario-based learning,'' in {\em Interactive Mobile Communication, Technologies and Learning}, pp.~106--113, Springer, 2023.

\bibitem{77}
T.~Shan and K.~Michel, ``Generative {AI} with {GOAP} for fast-paced dynamic decision-making in game environments,'' in {\em 2024 IEEE Conference on Games (CoG)}, pp.~1--5, 2024.

\bibitem{129}
T.~Rist, ``Using a large language model to turn explorations of virtual 3{D}-worlds into interactive narrative experiences,'' in {\em 2024 IEEE Conference on Games (CoG)}, pp.~1--8, 2024.

\bibitem{71}
C.-H. Chen and C.-L. Chang, ``Effectiveness of {AI}-assisted game-based learning on science learning outcomes, intrinsic motivation, cognitive load, and learning behavior,'' {\em Education and Information Technologies}, pp.~1--22, 2024.

\bibitem{96}
A.~Goslen, Y.~J. Kim, J.~Rowe, and J.~Lester, ``{LLM}-based student plan generation for adaptive scaffolding in game-based learning environments,'' {\em International Journal of Artificial Intelligence in Education}, pp.~1--26, 2024.

\bibitem{119}
G.~A. Molina-Barron, R.~E. Alvarado-Ramirez, and A.~Aguirre-Acosta, ``Storytelling: Digital narration enhanced by artificial intelligence in the metaverse,'' in {\em Proceedings of the 2023 7th International Conference on Education and E-Learning}, ICEEL '23, (New York, NY, USA), p.~33–40, Association for Computing Machinery, 2024.

\bibitem{60}
I.~Capecchi, T.~Borghini, and I.~Bernetti, ``{ARF}ood: Pioneering nutrition education for generation alpha through augmented reality and {AI}-driven serious gaming,'' in {\em International Conference on Extended Reality}, pp.~351--359, Springer, 2024.

\bibitem{83}
A.~Rychert, M.~L. Ganuza, and M.~N. Selzer, ``Integrating {GPT} as an assistant for low-cost virtual reality escape-room games,'' {\em IEEE Computer Graphics and Applications}, vol.~44, no.~4, pp.~14--25, 2024.

\bibitem{152}
E.~Whitaker, E.~Trewhitt, and E.~Veinott, ``Heuristica {II}: Updating a 2011 game-based training architecture using generative {AI} tools,'' in {\em International Conference on Human-Computer Interaction}, pp.~314--332, Springer, 2024.

\bibitem{175}
M.~Ribeiro, J.~Santos, J.~a. Lobo, S.~Ara\'{u}jo, L.~Magalh\~{a}es, and T.~Ad\~{a}o, ``{VR}, {AR}, gamification and {AI} towards the next generation of systems supporting cultural heritage: addressing challenges of a museum context,'' in {\em Proceedings of the 29th International ACM Conference on 3D Web Technology}, Web3D '24, (New York, NY, USA), Association for Computing Machinery, 2024.

\bibitem{148}
B.~W. Wilson, S.~Eswaran, O.~Riyaz, J.~Chiyah-Garcia, and M.~P. Aylett, ``Follow the yellow or red brick road? investigating the impact of narratives in a guided navigation task,'' in {\em Proceedings of the 12th International Conference on Human-Agent Interaction}, HAI '24, (New York, NY, USA), p.~411–413, Association for Computing Machinery, 2024.

\bibitem{154}
M.~Sidji, W.~Smith, and M.~J. Rogerson, ``Human-{AI} collaboration in cooperative games: A study of playing codenames with an {LLM} assistant,'' {\em Proc. ACM Hum.-Comput. Interact.}, vol.~8, Oct. 2024.

\bibitem{170}
M.~Shah, M.~Pankiewicz, R.~S. Baker, J.~Chi, Y.~Xin, H.~Shah, and D.~Fonseca, ``Students' use of an {LLM}-powered virtual teaching assistant for recommending educational applications of games,'' in {\em Serious Games} (J.~L. Plass and X.~Ochoa, eds.), (Cham), pp.~19--24, Springer Nature Switzerland, 2025.

\bibitem{143}
L.~Patra, M.~Kumari, and R.~Gopalapillai, ``Enhancing artificial intelligence and machine learning understanding through envision: A virtual reality approach,'' in {\em 2024 First International Conference on Pioneering Developments in Computer Science \& Digital Technologies (IC2SDT)}, pp.~516--521, 2024.

\bibitem{174}
X.~Qin and G.~Weaver, ``Utilizing generative {AI} for {VR} exploration testing: A case study,'' in {\em Proceedings of the 39th IEEE/ACM International Conference on Automated Software Engineering Workshops}, ASEW '24, (New York, NY, USA), p.~228–232, Association for Computing Machinery, 2024.

\bibitem{162}
K.~Plupattanakit, P.~Suntichaikul, P.~Taveekitworachai, R.~Thawonmas, J.~White, K.~Sookhanaphibarn, and W.~Choensawat, ``Llms in eduverse: {LLM}-integrated english educational game in metaverse,'' in {\em 2024 IEEE 13th Global Conference on Consumer Electronics (GCCE)}, pp.~257--258, 2024.

\bibitem{53}
J.~Kelly, M.~Mateas, and N.~Wardrip-Fruin, ``Towards computational support with language models for {TTRPG} game masters,'' in {\em Proceedings of the 18th International Conference on the Foundations of Digital Games}, FDG '23, (New York, NY, USA), Association for Computing Machinery, 2023.

\bibitem{70}
X.~You, P.~Taveekitworachai, S.~Chen, M.~Can~Gursesli, X.~Li, Y.~Xia, and R.~Thawonmas, ``Dungeons, dragons, and emotions: A preliminary study of player sentiment in {LLM}-driven ttrpgs,'' in {\em Proceedings of the 19th International Conference on the Foundations of Digital Games}, FDG '24, (New York, NY, USA), Association for Computing Machinery, 2024.

\bibitem{4}
K.~Frans, ``{AI} charades: Language models as interactive game environments,'' in {\em 2021 IEEE Conference on Games (CoG)}, pp.~1--2, IEEE, 2021.

\bibitem{3}
Y.~Yao, H.~Zhong, Z.~Zhang, X.~Han, X.~Wang, K.~Zhang, C.~Xiao, G.~Zeng, Z.~Liu, and M.~Sun, ``Adversarial language games for advanced natural language intelligence,'' in {\em Proceedings of the AAAI Conference on Artificial Intelligence}, vol.~35, pp.~14248--14256, 2021.

\bibitem{55}
C.~Jaramillo, M.~Charity, R.~Canaan, and J.~Togelius, ``Word autobots: Using transformers for word association in the game codenames,'' {\em Proceedings of the AAAI Conference on Artificial Intelligence and Interactive Digital Entertainment}, vol.~16, pp.~231--237, Oct. 2020.

\bibitem{13}
E.-y. Lee, N.~G.~D. il, G.-h. An, S.~Lee, and K.~Lim, ``{ChatGPT}-based debate game application utilizing prompt engineering,'' in {\em Proceedings of the 2023 International Conference on Research in Adaptive and Convergent Systems}, pp.~1--6, 2023.

\bibitem{42}
T.~S. Wang and A.~S. Gordon, ``Playing story creation games with large language models: Experiments with {GPT}-3.5,'' in {\em Interactive Storytelling} (L.~Holloway-Attaway and J.~T. Murray, eds.), (Cham), pp.~297--305, Springer Nature Switzerland, 2023.

\bibitem{112}
M.~Sidji and M.~Stephenson, ``Prompt engineering {C}hat{GPT} for codenames,'' in {\em 2024 IEEE Conference on Games (CoG)}, pp.~1--4, 2024.

\bibitem{125}
Q.~Ni, Y.~Yu, Y.~Ma, X.~Lin, C.~Deng, T.~Wei, and M.~Xuan, ``The social cognition ability evaluation of llms: A dynamic gamified assessment and hierarchical social learning measurement approach,'' {\em ACM Trans. Intell. Syst. Technol.}, June 2024.
\newblock Just Accepted.

\bibitem{108}
D.~Jeurissen, D.~Perez-Liebana, J.~Gow, D.~Çakmak, and J.~Kwan, ``Playing net{H}ack with {LLM}s: Potential \& limitations as zero-shot agents,'' in {\em 2024 IEEE Conference on Games (CoG)}, pp.~1--8, 2024.

\bibitem{80}
M.~Kim, ``Gpts in mafia-like game simulation,'' in {\em Extended Abstracts of the 2024 CHI Conference on Human Factors in Computing Systems}, CHI EA '24, (New York, NY, USA), Association for Computing Machinery, 2024.

\bibitem{82}
Z.~Shi, M.~Fang, L.~Chen, Y.~Du, and J.~Wang, ``Human-guided moral decision making in text-based games,'' {\em Proceedings of the AAAI Conference on Artificial Intelligence}, vol.~38, pp.~21574--21582, Mar. 2024.

\bibitem{153}
M.~Lamparth, A.~Corso, J.~Ganz, O.~S. Mastro, J.~Schneider, and H.~Trinkunas, ``Human vs. machine: Behavioral differences between expert humans and language models in wargame simulations,'' {\em Proceedings of the AAAI/ACM Conference on AI, Ethics, and Society}, vol.~7, pp.~807--817, Oct. 2024.

\bibitem{161}
S.~Zhong, Z.~Huang, S.~Gao, W.~Wen, L.~Lin, M.~Zitnik, and P.~Zhou, ``Let's think outside the box: Exploring leap-of-thought in large language models with creative humor generation,'' in {\em 2024 IEEE/CVF Conference on Computer Vision and Pattern Recognition (CVPR)}, pp.~13246--13257, 2024.

\bibitem{52}
M.~Ciolino, J.~Kalin, and D.~Noever, ``The {G}o transformer: Natural language modeling for game play,'' in {\em 2020 Third International Conference on Artificial Intelligence for Industries (AI4I)}, pp.~23--26, 2020.

\bibitem{131}
L.~Lawley and C.~Maclellan, ``Val: Interactive task learning with {GPT} dialog parsing,'' in {\em Proceedings of the 2024 CHI Conference on Human Factors in Computing Systems}, CHI '24, (New York, NY, USA), Association for Computing Machinery, 2024.

\bibitem{90}
B.~Bateni and J.~Whitehead, ``Language-driven play: Large language models as game-playing agents in slay the spire,'' in {\em Proceedings of the 19th International Conference on the Foundations of Digital Games}, FDG '24, (New York, NY, USA), Association for Computing Machinery, 2024.

\bibitem{87}
S.~Oh, I.~Chung, and K.-J. Kim, ``Lang{B}irds: An agent for angry birds using a large language model,'' in {\em 2024 IEEE Conference on Games (CoG)}, pp.~1--8, 2024.

\bibitem{139}
A.~L. Davis and G.~Sukthankar, ``Decoding chess mastery: A mechanistic analysis of a chess language transformer model,'' in {\em International Conference on Artificial General Intelligence}, pp.~63--72, Springer, 2024.

\bibitem{160}
Y.~Chen and S.~Chu, ``Large language models in wargaming: Methodology, application, and robustness,'' in {\em 2024 IEEE/CVF Conference on Computer Vision and Pattern Recognition Workshops (CVPRW)}, pp.~2894--2903, 2024.

\bibitem{178}
A.~de~Wynter, ``Will {GPT}-4 run doom?,'' {\em IEEE Transactions on Games}, pp.~1--10, 2024.

\bibitem{84}
Y.~Chen, S.~She, and Y.~Sun, ``Is {AI}-generated content better? a study based on ant forest game content recommendation,'' in {\em Proceedings of the 2024 Guangdong-Hong Kong-Macao Greater Bay Area International Conference on Digital Economy and Artificial Intelligence}, DEAI '24, (New York, NY, USA), p.~749–755, Association for Computing Machinery, 2024.

\bibitem{29}
M.~Lankes and A.~St{\"o}ckl, ``Game reviews reviewed: A game designer’s perspective on {AI}-generated game review analyses,'' in {\em 2023 IEEE Conference on Games (CoG)}, pp.~1--8, IEEE, 2023.

\bibitem{6}
X.~Li, X.~You, S.~Chen, P.~Taveekitworachai, and R.~Thawonmas, ``Analyzing audience comments: Improving interactive narrative with {ChatGPT},'' in {\em International Conference on Interactive Digital Storytelling}, pp.~220--228, Springer, 2023.

\bibitem{171}
U.~B.~R. Khan, N.~Akhtar, A.~Naaz, M.~Akram, and A.~Haque, ``Toward effective suggestion mining in game reviews: Introducing an aspect-based multiclass multilevel dataset,'' in {\em Doctoral Symposium on Computational Intelligence}, pp.~105--119, Springer, 2024.

\bibitem{68}
D.~Carlander, K.~Okada, H.~Engström, and S.~Kurabayashi, ``Controlled chain of thought: Eliciting role-play understanding in {LLM} through prompts,'' in {\em 2024 IEEE Conference on Games (CoG)}, pp.~1--4, 2024.

\bibitem{85}
A.~N. Tak and J.~Gratch, ``Is {GPT} a computational model of emotion?,'' in {\em 2023 11th International Conference on Affective Computing and Intelligent Interaction (ACII)}, pp.~1--8, 2023.

\bibitem{137}
G.~Barbero, M.~M{\"u}ller-Brockhausen, and M.~Preuss, ``Challenges of open world games for {AI}: Insights from human gameplay,'' in {\em Data Science and Artificial Intelligence} (C.~Anutariya, M.~M. Bonsangue, E.~Budhiarti-Nababan, and O.~S. Sitompul, eds.), (Singapore), pp.~127--141, Springer Nature Singapore, 2025.

\bibitem{95}
J.~Liao, S.~Li, Z.~Yang, J.~Wu, Y.~Yuan, X.~Wang, and X.~He, ``L{L}a{RA}: Large language-recommendation assistant,'' in {\em Proceedings of the 47th International ACM SIGIR Conference on Research and Development in Information Retrieval}, SIGIR '24, (New York, NY, USA), p.~1785–1795, Association for Computing Machinery, 2024.

\bibitem{156}
A.~Boz, W.~Zorgdrager, Z.~Kotti, J.~Harte, P.~Louridas, V.~Karakoidas, D.~Jannach, and M.~Fragkoulis, ``Improving sequential recommendations with {LLM}s,'' {\em ACM Trans. Recomm. Syst.}, Jan. 2025.
\newblock Just Accepted.

\bibitem{166}
J.~Zhang, R.~Xie, Y.~Hou, X.~Zhao, L.~Lin, and J.-R. Wen, ``Recommendation as instruction following: A large language model empowered recommendation approach,'' {\em ACM Trans. Inf. Syst.}, Dec. 2024.
\newblock Just Accepted.

\bibitem{113}
P.~Taveekitworachai, F.~Abdullah, M.~C. Gursesli, A.~Lanata, A.~Guazzini, and R.~Thawonmas, ``Prompt evolution through examples for large language models–a case study in game comment toxicity classification,'' in {\em 2024 IEEE International Workshop on Metrology for Industry 4.0 \& IoT (MetroInd4.0 \& IoT)}, pp.~22--27, 2024.

\bibitem{32}
V.~M. Carvalho and M.~A.~F. Rodrigues, ``Investigating and comparing the perceptions of voice interaction in digital games: Opportunities for health and wellness applications,'' in {\em 2023 IEEE 11th International Conference on Serious Games and Applications for Health (SeGAH)}, pp.~1--8, IEEE, 2023.

\bibitem{157}
X.~Liu, A.~Zambrano, A.~Barany, J.~Ocumpaugh, J.~Ginger, M.~Gadbury, H.~C. Lane, and R.~S. Baker, ``Investigating learner interest and observation patterns in a minecraft virtual astronomy environment,'' in {\em International Conference on Quantitative Ethnography}, pp.~19--34, Springer, 2024.

\end{thebibliography}


\begin{thebibliography}{100}

\bibitem{14}
F.~Gao, K.~Fang, and W.~K.~V. Chan, ``Chemical life: Knowledge-based personality, emotion and action cues in educational games,'' in {\em 2023 IEEE Conference on Games (CoG)}, pp.~1--3, IEEE, 2023.

\bibitem{7}
P.~Ammanabrolu, W.~Cheung, D.~Tu, W.~Broniec, and M.~Riedl, ``Bringing stories alive: Generating interactive fiction worlds,'' in {\em Proceedings of the AAAI Conference on Artificial Intelligence and Interactive Digital Entertainment}, vol.~16, pp.~3--9, 2020.

\bibitem{37}
M.~G. Torii, T.~Murakami, and Y.~Ochiai, ``Lottery and sprint: Generate a board game with design sprint method on {A}uto{GPT},'' in {\em Companion Proceedings of the Annual Symposium on Computer-Human Interaction in Play}, pp.~259--265, 2023.

\bibitem{93}
A.~Alkhayat, B.~Ciranni, R.~S. Tumuluri, and R.~S. Tulasi, ``Leveraging large language models for enhanced vr development: Insights and challenges,'' in {\em 2024 IEEE Gaming, Entertainment, and Media Conference (GEM)}, pp.~1--6, 2024.

\bibitem{102}
P.~Babiuch, A.~Łapczyński, H.~Jegierski, M.~Jegierski, R.~Salamon, B.~Kolber-Bugajska, M.~Płaza, S.~Deniziak, P.~Pięta, G.~Łukawski, A.~Jasiński, J.~Opałka, A.~Marmon, K.~Kwiatkowski, A.~Cybulski, M.~Igras-Cybulska, and P.~Węgrzyn, ``{MUN}: an {AI}-powered multiplayer networking solution for {VR} games,'' in {\em 2024 IEEE Conference on Virtual Reality and 3D User Interfaces Abstracts and Workshops (VRW)}, pp.~1204--1205, 2024.

\bibitem{35}
Y.~Sun, Z.~Li, K.~Fang, C.~H. Lee, and A.~Asadipour, ``Language as reality: a co-creative storytelling game experience in 1001 nights using generative {AI},'' in {\em Proceedings of the AAAI Conference on Artificial Intelligence and Interactive Digital Entertainment}, vol.~19, pp.~425--434, 2023.

\bibitem{10}
A.~Zhu, L.~Martin, A.~Head, and C.~Callison-Burch, ``{CALYPSO}: {LLM}s as dungeon master's assistants,'' in {\em Proceedings of the AAAI Conference on Artificial Intelligence and Interactive Digital Entertainment}, vol.~19, pp.~380--390, 2023.

\bibitem{122}
F.~Abdullah, P.~Taveekitworachai, M.~F. Dewantoro, R.~Thawonmas, J.~Togelius, and J.~Renz, ``The 1st {C}hat{GPT4PCG} competition,'' {\em IEEE Transactions on Games}, pp.~1--17, 2024.

\bibitem{147}
B.~Yoo and K.-J. Kim, ``Finding deceivers in social context with large language models and how to find them: the case of the mafia game,'' {\em Scientific Reports}, vol.~14, no.~1, p.~30946, 2024.

\bibitem{149}
D.~Li, S.~S. Sohn, S.~Zhang, C.-J. Chang, and M.~Kapadia, ``From words to worlds: Transforming one-line prompts into multi-modal digital stories with {LLM} agents,'' in {\em Proceedings of the 17th ACM SIGGRAPH Conference on Motion, Interaction, and Games}, MIG '24, (New York, NY, USA), Association for Computing Machinery, 2024.

\bibitem{98}
F.~De~La~Torre, C.~M. Fang, H.~Huang, A.~Banburski-Fahey, J.~Amores~Fernandez, and J.~Lanier, ``{LLMR}: Real-time prompting of interactive worlds using large language models,'' in {\em Proceedings of the 2024 CHI Conference on Human Factors in Computing Systems}, CHI '24, (New York, NY, USA), Association for Computing Machinery, 2024.

\bibitem{34}
P.~Taveekitworachai, M.~C. Gursesli, F.~Abdullah, S.~Chen, F.~Cala, A.~Guazzini, A.~Lanata, and R.~Thawonmas, ``Journey of {ChatGPT} from prompts to stories in games: the positive, the negative, and the neutral,'' in {\em 2023 IEEE 13th International Conference on Consumer Electronics-Berlin (ICCE-Berlin)}, pp.~202--203, IEEE, 2023.

\bibitem{38}
Q.~Kuang, F.~Shen, C.~M. Fang, and A.~Dong, ``Memeopoly: An {AI}-powered physical board game interface for tangible play and learning art and design,'' in {\em Companion Proceedings of the Annual Symposium on Computer-Human Interaction in Play}, pp.~292--297, 2023.

\bibitem{15}
R.~LC and V.~Ruijters, ``Chikyuchi: In-person/remote game exhibition for climate change influence,'' in {\em Sixteenth International Conference on Tangible, Embedded, and Embodied Interaction}, pp.~1--4, 2022.

\bibitem{46}
B.~Thabet, N.~Zanichelli, and F.~Zanichelli, ``Q{\&}{A} generation for flashcards within a transformer-based framework,'' in {\em Higher Education Learning Methodologies and Technologies Online} (G.~Fulantelli, D.~Burgos, G.~Casalino, M.~Cimitile, G.~Lo~Bosco, and D.~Taibi, eds.), (Cham), pp.~789--806, Springer Nature Switzerland, 2023.

\bibitem{23}
Y.~Mori and Y.~Miyake, ``Ethical issues in automatic dialogue generation for non-player characters in digital games,'' in {\em 2022 IEEE International Conference on Big Data (Big Data)}, pp.~5132--5139, IEEE, 2022.

\bibitem{28}
Q.~R. Yong and A.~Mitchell, ``From playing the story to gaming the system: Repeat experiences of a large language model-based interactive story,'' in {\em International Conference on Interactive Digital Storytelling}, pp.~395--409, Springer, 2023.

\bibitem{22}
W.~Hettmann, M.~W{\"o}lfel, M.~Butz, K.~Torner, and J.~Finken, ``Engaging museum visitors with {AI}-generated narration and gameplay,'' in {\em International Conference on ArtsIT, Interactivity and Game Creation}, pp.~201--214, Springer, 2022.

\bibitem{54}
P.~Taveekitworachai, F.~Abdullah, M.~C. Gursesli, M.~F. Dewantoro, S.~Chen, A.~Lanata, A.~Guazzini, and R.~Thawonmas, ``What is waiting for us at the end? inherent biases of game story endings in large language models,'' in {\em Interactive Storytelling} (L.~Holloway-Attaway and J.~T. Murray, eds.), (Cham), pp.~274--284, Springer Nature Switzerland, 2023.

\bibitem{20}
K.~Saito, K.~Kobayashi, W.~Takekoshi, A.~Hashimoto, N.~Hirai, A.~Kimura, A.~Takahashi, N.~Yoshioka, and A.~Mano, ``Double impact: Children’s serious {RPG} generation/play with a large language model for their deeper engagement in social issues,'' in {\em Joint International Conference on Serious Games}, pp.~274--289, Springer, 2023.

\bibitem{25}
T.~Triyason, ``Exploring the potential of {ChatGPT} as a dungeon master in dungeons \& dragons tabletop game,'' in {\em Proceedings of the 13th International Conference on Advances in Information Technology}, pp.~1--6, 2023.

\bibitem{26}
C.~Ang, L.~R. Cortel, C.~L. Santos, and E.~Ong, ``Fable reborn: Investigating gameplay experience between a human player and a virtual dungeon master,'' in {\em Extended Abstracts of the 2023 CHI Conference on Human Factors in Computing Systems}, pp.~1--7, 2023.

\bibitem{57}
H.~Tang and M.~Singha, ``A mystery for you: A fact-checking game enhanced by large language models ({LLMs}) and a tangible interface,'' in {\em Extended Abstracts of the 2024 CHI Conference on Human Factors in Computing Systems}, CHI EA '24, (New York, NY, USA), Association for Computing Machinery, 2024.

\bibitem{58}
P.~Mieschke and S.~Radicke, ``Adaptive {LLM}-based game radio ({ALGR}),'' in {\em 2023 4th International Conference on Computers and Artificial Intelligence Technology (CAIT)}, pp.~277--283, 2023.

\bibitem{61}
P.~Taveekitworachai, K.~Plupattanakit, and R.~Thawonmas, ``Assessing inherent biases following prompt compression of large language models for game story generation,'' in {\em 2024 IEEE Conference on Games (CoG)}, pp.~1--4, 2024.

\bibitem{63}
W.~Du, Z.~Zhu, X.~Xu, H.~Che, and S.~Chen, ``Careersim: Gamification design leveraging {LLM}s for career development reflection,'' in {\em Extended Abstracts of the 2024 CHI Conference on Human Factors in Computing Systems}, CHI EA '24, (New York, NY, USA), Association for Computing Machinery, 2024.

\bibitem{69}
S.~R. Cox and W.~T. Ooi, ``Conversational interactions with npcs in {LLM}-driven gaming: Guidelines from a content analysis of player feedback,'' in {\em International Workshop on Chatbot Research and Design}, pp.~167--184, Springer, 2023.

\bibitem{86}
P.~Taveekitworachai, M.~C. Gursesli, F.~Abdullah, S.~Chen, F.~Cala, A.~Guazzini, A.~Lanata, and R.~Thawonmas, ``Journey of {C}hat{GPT} from prompts to stories in games: the positive, the negative, and the neutral,'' in {\em 2023 IEEE 13th International Conference on Consumer Electronics - Berlin (ICCE-Berlin)}, pp.~202--203, 2023.

\bibitem{92}
S.~Buongiorno and C.~Clark, ``Leveraging gaming to enhance knowledge graphs for explainable generative {AI} applications,'' in {\em 2024 IEEE Conference on Games (CoG)}, pp.~1--4, 2024.

\bibitem{123}
A.~Normoyle, S.~J\"{o}rg, and J.~Hill, ``The curation tree: A lightweight behavior tree framework for implementing puzzle and narrative games,'' in {\em Proceedings of the 19th International Conference on the Foundations of Digital Games}, FDG '24, (New York, NY, USA), Association for Computing Machinery, 2024.

\bibitem{91}
T.~Yu, M.~Chen, Y.~Li, D.~Lew, and K.~Yu, ``La{S}ofa: Integrating fantasy storytelling in human-robot interaction through an interactive sofa robot,'' in {\em Companion of the 2024 ACM/IEEE International Conference on Human-Robot Interaction}, HRI '24, (New York, NY, USA), p.~1168–1172, Association for Computing Machinery, 2024.

\bibitem{159}
R.~Farrell and S.~G. Ware, ``Large language models as narrative planning search guides,'' {\em IEEE Transactions on Games}, pp.~1--10, 2024.

\bibitem{135}
R.~Y. Camilleri and V.~Camilleri, ``Beyond the maze: How {AI} personalizes learning and drives engagement in educational games,'' in {\em Proceedings of the Future Technologies Conference (FTC) 2024, Volume 2} (K.~Arai, ed.), (Cham), pp.~301--320, Springer Nature Switzerland, 2024.

\bibitem{163}
T.~Merino, S.~Earle, R.~Sudhakaran, S.~Sudhakaran, and J.~Togelius, ``Making new connections: {LLM}s as puzzle generators for the new york times’ connections word game,'' {\em Proceedings of the AAAI Conference on Artificial Intelligence and Interactive Digital Entertainment}, vol.~20, pp.~87--96, Nov. 2024.

\bibitem{27}
J.~van Stegeren and J.~My{\'s}liwiec, ``Fine-tuning {GPT}-2 on annotated {RPG} quests for {NPC} dialogue generation,'' in {\em Proceedings of the 16th International Conference on the Foundations of Digital Games}, pp.~1--8, 2021.

\bibitem{41}
T.~Ashby, B.~K. Webb, G.~Knapp, J.~Searle, and N.~Fulda, ``Personalized quest and dialogue generation in role-playing games: A knowledge graph- and language model-based approach,'' in {\em Proceedings of the 2023 CHI Conference on Human Factors in Computing Systems}, CHI '23, (New York, NY, USA), Association for Computing Machinery, 2023.

\bibitem{47}
S.~Al-Nassar, A.~Schaap, M.~V.~D. Zwart, M.~Preuss, and M.~A. G\'{o}mez-Maureira, ``Questville: Procedural quest generation using {NLP} models,'' in {\em Proceedings of the 18th International Conference on the Foundations of Digital Games}, FDG '23, (New York, NY, USA), Association for Computing Machinery, 2023.

\bibitem{31}
S.~V{\"a}rtinen, P.~H{\"a}m{\"a}l{\"a}inen, and C.~Guckelsberger, ``Generating role-playing game quests with {GPT} language models,'' {\em IEEE Transactions on Games}, 2022.

\bibitem{45}
S.~Sudhakaran, M.~Gonz\'{a}lez-Duque, C.~Glanois, M.~Freiberger, E.~Najarro, and S.~Risi, ``Prompt-guided level generation,'' in {\em Proceedings of the Companion Conference on Genetic and Evolutionary Computation}, GECCO '23 Companion, (New York, NY, USA), p.~179–182, Association for Computing Machinery, 2023.

\bibitem{36}
G.~Todd, S.~Earle, M.~U. Nasir, M.~C. Green, and J.~Togelius, ``Level generation through large language models,'' in {\em Proceedings of the 18th International Conference on the Foundations of Digital Games}, FDG '23, (New York, NY, USA), Association for Computing Machinery, 2023.

\bibitem{43}
M.~U. Nasir and J.~Togelius, ``Practical {PCG} through large language models,'' in {\em 2023 IEEE Conference on Games (CoG)}, pp.~1--4, 2023.

\bibitem{56}
S.~Hu, Z.~Huang, C.~Hu, and J.~Liu, ``3d building generation in minecraft via large language models,'' in {\em 2024 IEEE Conference on Games (CoG)}, pp.~1--4, 2024.

\bibitem{132}
D.~Hafnar and J.~Demšar, ``Zero-shot reasoning: Personalized content generation without the cold start problem,'' {\em IEEE Transactions on Games}, pp.~1--10, 2024.

\bibitem{1}
T.-H. Fu and K.-C. Wu, ``A dynamic fitness game content generation system based on machine learning,'' in {\em Artificial Intelligence in HCI} (H.~Degen and S.~Ntoa, eds.), (Cham), pp.~50--62, Springer Nature Switzerland, 2023.

\bibitem{21}
J.-Y. Zhou and T.-H. Fu, ``Empowering interactive fitness game,'' in {\em 2023 International Conference on Consumer Electronics - Taiwan (ICCE-Taiwan)}, pp.~279--280, 2023.

\bibitem{101}
S.~Zheng, K.~He, L.~Yang, and J.~Xiong, ``Memory{R}epository for {AI} {NPC},'' {\em IEEE Access}, vol.~12, pp.~62581--62596, 2024.

\bibitem{130}
A.~Normoyle, J.~Sedoc, and F.~Durupinar, ``Using {LLM}s to animate interactive story characters with emotions and personality,'' in {\em 2024 IEEE Conference on Virtual Reality and 3D User Interfaces Abstracts and Workshops (VRW)}, pp.~632--635, 2024.

\bibitem{106}
X.~Peng, J.~Quaye, S.~Rao, W.~Xu, P.~Botchway, C.~Brockett, N.~Jojic, G.~DesGarennes, K.~Lobb, M.~Xu, J.~Leandro, C.~Jin, and B.~Dolan, ``Player-driven emergence in {LLM}-driven game narrative,'' in {\em 2024 IEEE Conference on Games (CoG)}, pp.~1--8, 2024.

\bibitem{73}
J.~Sissler, ``Enhancing non-player characters in {U}nity 3{D} using {GPT}-3.5,'' {\em ACM Games}, vol.~2, Aug. 2024.

\bibitem{88}
S.~Karaosmanoglu, E.~L. Fittschen, H.~Eyicalis, D.~Kraus, H.~Nickelmann, A.~Tomko, and F.~Steinicke, ``Language of {Z}elda: Facilitating language learning practices using {C}hat{GPT},'' in {\em Extended Abstracts of the 2024 CHI Conference on Human Factors in Computing Systems}, CHI EA '24, (New York, NY, USA), Association for Computing Machinery, 2024.

\bibitem{89}
Y.~Zhao, J.~Pan, Y.~Dong, T.~Dong, G.~Wang, F.~Ying, Q.~Shen, and J.~Cao, ``Language urban odyssey: A serious game for enhancing second language acquisition through large language models,'' in {\em Extended Abstracts of the 2024 CHI Conference on Human Factors in Computing Systems}, CHI EA '24, (New York, NY, USA), Association for Computing Machinery, 2024.

\bibitem{116}
A.~Y. Cheng, M.~Guo, M.~Ran, A.~Ranasaria, A.~Sharma, A.~Xie, K.~N. Le, B.~Vinaithirthan, S.~T. Luan, D.~T.~H. Wright, A.~Cuadra, R.~Pea, and J.~A. Landay, ``Scientific and fantastical: Creating immersive, culturally relevant learning experiences with augmented reality and large language models,'' in {\em Proceedings of the 2024 CHI Conference on Human Factors in Computing Systems}, CHI '24, (New York, NY, USA), Association for Computing Machinery, 2024.

\bibitem{127}
B.~Ngaw, G.~Jena, J.~a. Sedoc, and A.~Normoyle, ``Towards authoring open-ended behaviors for narrative puzzle games with large language model support,'' in {\em Proceedings of the 19th International Conference on the Foundations of Digital Games}, FDG '24, (New York, NY, USA), Association for Computing Machinery, 2024.

\bibitem{99}
M.~Dai, C.~Yuan, and X.~Nie, ``Managing the personality of {NPC}s with your interactions: A game design system based on large language models,'' in {\em International Conference on Human-Computer Interaction}, pp.~247--259, Springer, 2024.

\bibitem{105}
J.~Kelly, M.~Mateas, and N.~Wardrip-Fruin, ``Paradise: An experiment extending the ensemble social physics engine with language models,'' in {\em Proceedings of the 19th International Conference on the Foundations of Digital Games}, FDG '24, (New York, NY, USA), Association for Computing Machinery, 2024.

\bibitem{121}
Q.~Sun, Q.~Luo, Y.~Ni, and H.~Mi, ``Text2{AC}: A framework for game-ready 2{D} agent character({AC}) generation from natural language,'' in {\em Extended Abstracts of the 2024 CHI Conference on Human Factors in Computing Systems}, CHI EA '24, (New York, NY, USA), Association for Computing Machinery, 2024.

\bibitem{104}
R.~Zhao, W.~Zhang, J.~Li, L.~Zhu, Y.~Li, Y.~He, and L.~Gui, ``Narrative{P}lay: An automated system for crafting visual worlds in novels for role-playing,'' {\em Proceedings of the AAAI Conference on Artificial Intelligence}, vol.~38, pp.~23859--23861, Mar. 2024.

\bibitem{133}
M.~C. Uludagli and K.~Oguz, ``A social network generator for games evaluated against a real {NPC} network with {GPT}-generated node attributes,'' in {\em 2024 Innovations in Intelligent Systems and Applications Conference (ASYU)}, pp.~1--5, 2024.

\bibitem{145}
L.~J. Klinkert, S.~Buongiorno, and C.~Clark, ``Evaluating the efficacy of {LLM}s to emulate realistic human personalities,'' {\em Proceedings of the AAAI Conference on Artificial Intelligence and Interactive Digital Entertainment}, vol.~20, pp.~65--75, Nov. 2024.

\bibitem{165}
C.~Gr{\'e}visse, ``Ras{P}atient pi: A low-cost customizable {LLM}-based virtual standardized patient simulator,'' in {\em Applied Informatics} (H.~Florez and H.~Astudillo, eds.), (Cham), pp.~125--137, Springer Nature Switzerland, 2025.

\bibitem{142}
K.~Tsuyuguchi, K.~Shimizu, and K.~Suzuki, ``Emotion overflow: an interactive system to represent emotion with fluid,'' in {\em Adjunct Proceedings of the 37th Annual ACM Symposium on User Interface Software and Technology}, UIST Adjunct '24, (New York, NY, USA), Association for Computing Machinery, 2024.

\bibitem{167}
S.~S. Maram, Y.~Malegaonkar, M.~Escarce~Junior, and M.~Seif El-Nasr, ``Shloka: Developing climate change interventions through a lens of religion and videogames,'' in {\em Companion Proceedings of the 2024 Annual Symposium on Computer-Human Interaction in Play}, CHI PLAY Companion '24, (New York, NY, USA), p.~181–187, Association for Computing Machinery, 2024.

\bibitem{17}
L.~Ferreira, L.~Lelis, and J.~Whitehead, ``Computer-generated music for tabletop role-playing games,'' in {\em Proceedings of the AAAI Conference on Artificial Intelligence and Interactive Digital Entertainment}, vol.~16, pp.~59--65, 2020.

\bibitem{5}
C.~Nimpattanavong, P.~Taveekitworachai, I.~Khan, T.~V. Nguyen, R.~Thawonmas, W.~Choensawat, and K.~Sookhanaphibarn, ``Am {I} fighting well? fighting game commentary generation with {ChatGPT},'' in {\em Proceedings of the 13th International Conference on Advances in Information Technology}, pp.~1--7, 2023.

\bibitem{39}
R.~AlJammaz, M.~Mateas, and N.~Wardrip-Fruin, ``Modeling morality-based argumentation for believable game characters: a design postmortem,'' in {\em Proceedings of the AAAI Conference on Artificial Intelligence and Interactive Digital Entertainment}, vol.~19, pp.~185--194, 2023.

\bibitem{30}
F.~Horn, S.~Vogt, and S.~P. G{\"o}bel, ``Game{TUL}earn: An interactive educational game authoring tool for 3{D} environments,'' in {\em Joint International Conference on Serious Games}, pp.~384--390, Springer, 2023.

\bibitem{49}
V.~Kumaran, J.~Rowe, B.~Mott, and J.~Lester, ``Scene{C}raft: Automating interactive narrative scene generation in digital games with large language models,'' {\em Proceedings of the AAAI Conference on Artificial Intelligence and Interactive Digital Entertainment}, vol.~19, pp.~86--96, Oct. 2023.

\bibitem{50}
L.~R. Kouzelis and O.~Spantidi, ``Synthesizing play-ready {VR} scenes with natural language prompts through {GPT} {API},'' in {\em Advances in Visual Computing} (G.~Bebis, G.~Ghiasi, Y.~Fang, A.~Sharf, Y.~Dong, C.~Weaver, Z.~Leo, J.~J. LaViola~Jr., and L.~Kohli, eds.), (Cham), pp.~15--26, Springer Nature Switzerland, 2023.

\bibitem{40}
V.~Burkus, A.~K{\'a}rp{\'a}ti, and L.~Sz{\'e}csi, ``{NLP}-assisted educational memory game experiment,'' in {\em Methodologies and Intelligent Systems for Technology Enhanced Learning, Workshops - 13th International Conference} (Z.~Kubincov{\'a}, F.~Caruso, T.-e. Kim, M.~Ivanova, L.~Lancia, and M.~A. Pellegrino, eds.), (Cham), pp.~59--69, Springer Nature Switzerland, 2023.

\bibitem{107}
B.~Feng, M.~Su, K.~Zeng, and X.~Li, ``Player-oriented procedural generation: Producing desired game content by natural language,'' in {\em International Conference on Human-Computer Interaction}, pp.~260--274, Springer, 2024.

\bibitem{110}
V.~Kumaran, D.~Carpenter, J.~Rowe, B.~Mott, and J.~Lester, ``Procedural level generation in educational games from natural language instruction,'' {\em IEEE Transactions on Games}, pp.~1--10, 2024.

\bibitem{97}
R.~Gallotta, A.~Liapis, and G.~Yannakakis, ``L{LM}aker: A game level design interface using (only) natural language,'' in {\em 2024 IEEE Conference on Games (CoG)}, pp.~1--2, 2024.

\bibitem{109}
X.~Zhang, F.~Wan, K.~Zhang, and H.~Jiang, ``Possible applications of language models in visual novel,'' in {\em 2023 International Conference on Educational Knowledge and Informatization (EKI)}, pp.~60--63, 2023.

\bibitem{67}
R.~Gallotta, A.~Liapis, and G.~Yannakakis, ``Consistent game content creation via function calling for large language models,'' in {\em 2024 IEEE Conference on Games (CoG)}, pp.~1--4, 2024.

\bibitem{75}
D.~Sezen, A.~Akcali, T.~I. Sezen, S.~Perks, and P.~Stewart, ``Exploring women's role in creative industries through collaborative action research using tabletop role-playing games,'' in {\em EAI International Conference on Technology, Innovation, Entrepreneurship and Education}, pp.~39--53, Springer, 2023.

\bibitem{78}
J.~Leandro, S.~Rao, M.~Xu, W.~Xu, N.~Jojic, C.~Brockett, and B.~Dolan, ``{GENEVA}: {GENE}rating and visualizing branching narratives using {LLM}s,'' in {\em 2024 IEEE Conference on Games (CoG)}, pp.~1--5, 2024.

\bibitem{94}
M.~Yin, E.~Wang, C.~Ng, and R.~Xiao, ``Lies, deceit, and hallucinations: Player perception and expectations regarding trust and deception in games,'' in {\em Proceedings of the 2024 CHI Conference on Human Factors in Computing Systems}, CHI '24, (New York, NY, USA), Association for Computing Machinery, 2024.

\bibitem{103}
J.~Li, Z.~Chen, W.~Lin, L.~Zou, X.~Xie, Y.~Hu, and D.~Li, ``Mystery game script compose based on a large language model,'' in {\em 2024 IEEE World AI IoT Congress (AIIoT)}, pp.~451--455, 2024.

\bibitem{134}
D.-A. Ciungan, N.~Goga, I.-A. Bratosin, and R.-C. Popa, ``An intelligent system for image generation in unity,'' in {\em 2024 IEEE SmartBlock4Africa}, pp.~1--5, 2024.

\bibitem{168}
L.~Ling, X.~Chen, R.~Wen, T.~J.-J. Li, and R.~LC, ``Sketchar: Supporting character design and illustration prototyping using generative {AI},'' {\em Proc. ACM Hum.-Comput. Interact.}, vol.~8, Oct. 2024.

\bibitem{169}
E.~M. Yi~Chan, C.~K. Seow, E.~S. Wee~Tan, M.~Wang, P.~C. Yau, and Q.~Cao, ``Sketchboard: Sketch-guided storyboard generation for game characters in the game industry,'' in {\em 2024 IEEE 22nd International Conference on Industrial Informatics (INDIN)}, pp.~1--8, 2024.

\bibitem{173}
F.~Bonetti, A.~Bucchiarone, and V.~Wanick, ``Using {LLM}s to adapt serious games with educators in the loop,'' in {\em Games and Learning Alliance} (A.~Sch{\"o}nbohm, F.~Bellotti, A.~Bucchiarone, F.~de~Rosa, M.~Ninaus, A.~Wang, V.~Wanick, and P.~Dondio, eds.), (Cham), pp.~68--77, Springer Nature Switzerland, 2025.

\bibitem{136}
Z.~Wu and H.~Sato, ``Build-it-here: Utilizing 2{D} inpainting models for on-site 3{D} object generation,'' in {\em 2024 IEEE 9th International Conference on Computational Intelligence and Applications (ICCIA)}, pp.~104--108, 2024.

\bibitem{176}
L.~Zhang, J.~Pan, J.~Gettig, S.~Oney, and A.~Guo, ``{VRC}opilot: Authoring 3d layouts with generative {AI} models in {VR},'' in {\em Proceedings of the 37th Annual ACM Symposium on User Interface Software and Technology}, UIST '24, (New York, NY, USA), Association for Computing Machinery, 2024.

\bibitem{164}
S.~Buongiorno, L.~Klinkert, Z.~Zhuang, T.~Chawla, and C.~Clark, ``{PANG}e{A}: Procedural artificial narrative using generative {AI} for turn-based, role-playing video games,'' {\em Proceedings of the AAAI Conference on Artificial Intelligence and Interactive Digital Entertainment}, vol.~20, pp.~156--166, Nov. 2024.

\bibitem{2}
M.~Charity, Y.~Bhartia, D.~Zhang, A.~Khalifa, and J.~Togelius, ``A preliminary study on a conceptual game feature generation and recommendation system,'' {\em arXiv preprint arXiv:2308.13538}, 2023.

\bibitem{33}
T.~Fulcini and M.~Torchiano, ``Is {ChatGPT} capable of crafting gamification strategies for software engineering tasks?,'' in {\em Proceedings of the 2nd International Workshop on Gamification in Software Development, Verification, and Validation}, pp.~22--28, 2023.

\bibitem{11}
P.~L. Lanzi and D.~Loiacono, ``{C}hat{GPT} and other large language models as evolutionary engines for online interactive collaborative game design,'' GECCO '23, p.~1383–1390, ACM, 2023.

\bibitem{12}
A.~M. Grow and F.~Khosmood, ``{ChatGPT} gamejam: Unleashing the power of large language models for game jams,'' in {\em Proceedings of the 7th International Conference on Game Jams, Hackathons and Game Creation Events}, pp.~51--54, 2023.

\bibitem{76}
C.~Hu, Y.~Zhao, and J.~Liu, ``Game generation via large language models,'' in {\em 2024 IEEE Conference on Games (CoG)}, pp.~1--4, 2024.

\bibitem{124}
A.~Anjum, Y.~Li, N.~Law, M.~Charity, and J.~Togelius, ``The ink splotch effect: A case study on {C}hat{GPT} as a co-creative game designer,'' in {\em Proceedings of the 19th International Conference on the Foundations of Digital Games}, FDG '24, (New York, NY, USA), Association for Computing Machinery, 2024.

\bibitem{151}
T.~Tanaka and E.~Simo-Serra, ``Grammar-based game description generation using large language models,'' {\em IEEE Transactions on Games}, pp.~1--14, 2024.

\bibitem{51}
T.~Merino, R.~Negri, D.~Rajesh, M.~Charity, and J.~Togelius, ``The five-dollar model: Generating game maps and sprites from sentence embeddings,'' {\em Proceedings of the AAAI Conference on Artificial Intelligence and Interactive Digital Entertainment}, vol.~19, pp.~107--115, Oct. 2023.

\bibitem{79}
M.~R. Taesiri, T.~Feng, C.-P. Bezemer, and A.~Nguyen, ``Glitch{B}ench: Can large multimodal models detect video game glitches?,'' in {\em 2024 IEEE/CVF Conference on Computer Vision and Pattern Recognition (CVPR)}, pp.~22444--22455, 2024.

\bibitem{59}
I.~J. Pérez-Colado, M.~Freire-Morán, A.~Calvo-Morata, V.~M. Pérez-Colado, and B.~Fernández-Manjón, ``Ai asyet another tool in undergraduate student projects: Preliminary results,'' in {\em 2024 IEEE Global Engineering Education Conference (EDUCON)}, pp.~1--7, 2024.

\bibitem{65}
I.-C. Baek, T.-H. Park, J.-H. Noh, C.-M. Bae, and K.-J. Kim, ``Chat{PCG}: Large language model-driven reward design for procedural content generation,'' in {\em 2024 IEEE Conference on Games (CoG)}, pp.~1--4, 2024.

\bibitem{128}
L.~Bingli and D.~V. Vargas, ``Towards immersive computational storytelling: Card-framework for enhanced persona-driven dialogues,'' {\em IEEE Transactions on Games}, pp.~1--13, 2024.

\bibitem{172}
P.~Wightman, ``Twisted games: A first experience of inclusion of {AI} tools in first year programming classes,'' in {\em 2024 IEEE Latin American Conference on Computational Intelligence (LA-CCI)}, pp.~1--6, 2024.

\bibitem{9}
H.~Osone, J.-L. Lu, and Y.~Ochiai, ``Buncho: {AI} supported story co-creation via unsupervised multitask learning to increase writers’ creativity in {J}apanese,'' in {\em Extended Abstracts of the 2021 CHI Conference on Human Factors in Computing Systems}, pp.~1--10, 2021.

\bibitem{44}
J.~Freiknecht and W.~Effelsberg, ``Procedural generation of interactive stories using language models,'' in {\em Proceedings of the 15th International Conference on the Foundations of Digital Games}, FDG '20, (New York, NY, USA), Association for Computing Machinery, 2020.

\bibitem{16}
E.~Nichols, L.~Gao, and R.~Gomez, ``Collaborative storytelling with large-scale neural language models,'' in {\em Proceedings of the 13th ACM SIGGRAPH Conference on Motion, Interaction and Games}, pp.~1--10, 2020.

\bibitem{19}
M.~Elgarf, S.~Zojaji, G.~Skantze, and C.~Peters, ``Creativebot: a creative storyteller robot to stimulate creativity in children,'' in {\em Proceedings of the 2022 International Conference on Multimodal Interaction}, pp.~540--548, 2022.

\bibitem{48}
H.~Shakeri, C.~Neustaedter, and S.~DiPaola, ``{SAGA}: Collaborative storytelling with {GPT}-3,'' in {\em Companion Publication of the 2021 Conference on Computer Supported Cooperative Work and Social Computing}, CSCW '21 Companion, (New York, NY, USA), p.~163–166, Association for Computing Machinery, 2021.

\bibitem{24}
B.~M. McLaren, ``Evaluating {{ChatGPT}}’s decimal skills and feedback generation in a digital learning game,'' in {\em Responsive and Sustainable Educational Futures: 18th European Conference on Technology Enhanced Learning, EC-TEL 2023, Aveiro, Portugal, September 4--8, 2023, Proceedings}, vol.~14200, p.~278, Springer Nature, 2023.

\bibitem{8}
Y.~Sun, X.~Ni, H.~Feng, R.~LC, C.~H. Lee, and A.~Asadipour, ``Bringing stories to life in 1001 nights: A co-creative text adventure game using a story generation model,'' in {\em International Conference on Interactive Digital Storytelling}, pp.~651--672, Springer, 2022.

\bibitem{126}
R.~Divanji, A.~Dangol, E.~J. Lombard, K.~Chen, and J.~D. Rubin, ``Togethertales {RPG}: Prosocial skill development through digitally mediated collaborative role-playing,'' in {\em Proceedings of the 23rd Annual ACM Interaction Design and Children Conference}, IDC '24, (New York, NY, USA), p.~1012–1015, Association for Computing Machinery, 2024.

\bibitem{74}
S.~Zhou, L.~B. Hendra, Q.~Zhang, J.~Holopainen, and R.~LC, ``Eternagram: Probing player attitudes towards climate change using a {C}hat{GPT}-driven text-based adventure,'' in {\em Proceedings of the 2024 CHI Conference on Human Factors in Computing Systems}, CHI '24, (New York, NY, USA), Association for Computing Machinery, 2024.

\bibitem{115}
P.~Almeida, A.~Teixeira, A.~Velhinho, R.~Raposo, T.~Silva, and L.~Pedro, ``Remixing and repurposing cultural heritage archives through a collaborative and {AI}-generated storytelling digital platform,'' in {\em Proceedings of the 2024 ACM International Conference on Interactive Media Experiences Workshops}, IMXw '24, (New York, NY, USA), p.~100–104, Association for Computing Machinery, 2024.

\bibitem{118}
D.~Yang, E.~Kleinman, G.~M. Troiano, E.~Tochilnikova, and C.~Harteveld, ``Snake story: Exploring game mechanics for mixed-initiative co-creative storytelling games,'' in {\em Proceedings of the 19th International Conference on the Foundations of Digital Games}, FDG '24, (New York, NY, USA), Association for Computing Machinery, 2024.

\bibitem{120}
Y.~Wang, Q.~Zhou, and D.~Ledo, ``Story{V}erse: Towards co-authoring dynamic plot with {LLM}-based character simulation via narrative planning,'' in {\em Proceedings of the 19th International Conference on the Foundations of Digital Games}, FDG '24, (New York, NY, USA), Association for Computing Machinery, 2024.

\bibitem{62}
B.~Lyman, A.~Ebrahimi, J.~Cox, S.~Chan, C.~Barney, and B.~De~Schutter, ``Cardistry: Exploring a gpt model workflow as an adapted method of gaminiscing,'' in {\em Proceedings of the 19th International Conference on the Foundations of Digital Games}, FDG '24, (New York, NY, USA), Association for Computing Machinery, 2024.

\bibitem{64}
E.~S. De~Lima, B.~Feij\'{o}, M.~A. Cassanova, and A.~L. Furtado, ``Chat{G}eppetto - an {AI}-powered storyteller,'' in {\em Proceedings of the 22nd Brazilian Symposium on Games and Digital Entertainment}, SBGames '23, (New York, NY, USA), p.~28–37, Association for Computing Machinery, 2024.

\bibitem{66}
C.~Lee and J.~R. Mindel, ``Closer and closer worlds: Using {LLM}s to surface personal stories in world-building conversation games,'' in {\em Companion Publication of the 2024 ACM Designing Interactive Systems Conference}, DIS '24 Companion, (New York, NY, USA), p.~289–293, Association for Computing Machinery, 2024.

\bibitem{72}
T.~Reichert, M.~Miftari, C.~Herling, and N.~Marsden, ``Empowering female founders with {AI} and play: Integration of a large language model into a serious game with player-generated content,'' in {\em International Conference on Human-Computer Interaction}, pp.~69--83, Springer, 2024.

\bibitem{100}
C.~Zhang, X.~Liu, K.~Ziska, S.~Jeon, C.-L. Yu, and Y.~Xu, ``Mathemyths: Leveraging large language models to teach mathematical language through child-{AI} co-creative storytelling,'' in {\em Proceedings of the 2024 CHI Conference on Human Factors in Computing Systems}, CHI '24, (New York, NY, USA), Association for Computing Machinery, 2024.

\bibitem{81}
M.~Valentim, J.~Falk, and N.~Inie, ``Hacc-man: An arcade game for jailbreaking llms,'' in {\em Companion Publication of the 2024 ACM Designing Interactive Systems Conference}, DIS '24 Companion, (New York, NY, USA), p.~338–341, Association for Computing Machinery, 2024.

\bibitem{138}
J.~P. Vieira~Sousa, P.~Campos, and P.~Bala, ``College tales: Pilot study on large language models generated narratives for mental health literacy,'' in {\em Proceedings of the 27th International Academic Mindtrek Conference}, Mindtrek '24, (New York, NY, USA), p.~270–275, Association for Computing Machinery, 2024.

\bibitem{140}
A.~B\"{o}r\"{u}tecene and O.~O. Buruk, ``Dr. ping and dr. pong: Rethinking writing and work with playful embodied ais,'' in {\em Proceedings of the Halfway to the Future Symposium}, HttF '24, (New York, NY, USA), Association for Computing Machinery, 2024.

\bibitem{158}
V.~Shrivastava, S.~Sharma, D.~Chakraborty, and M.~Kinnula, ``Is a sunny day bright and cheerful or hot and uncomfortable? young children's exploration of chat{GPT},'' in {\em Proceedings of the 13th Nordic Conference on Human-Computer Interaction}, NordiCHI '24, (New York, NY, USA), Association for Computing Machinery, 2024.

\bibitem{144}
T.~Tucek, ``Enhancing empathy through personalized {AI}-driven experiences and conversations with digital humans in video games,'' in {\em Companion Proceedings of the 2024 Annual Symposium on Computer-Human Interaction in Play}, CHI PLAY Companion '24, (New York, NY, USA), p.~446–449, Association for Computing Machinery, 2024.

\bibitem{141}
Y.~Lyu, D.~Liu, P.~An, X.~Tong, H.~Zhang, K.~Katsuragawa, and J.~Zhao, ``Emooly: Supporting autistic children in collaborative social-emotional learning with caregiver participation through interactive {AI}-infused and {AR} activities,'' {\em Proc. ACM Interact. Mob. Wearable Ubiquitous Technol.}, vol.~8, Nov. 2024.

\bibitem{146}
F.~R. Christiansen, L.~N. Hollensberg, N.~B. Jensen, K.~Julsgaard, K.~N. Jespersen, and I.~Nikolov, ``Exploring presence in interactions with {LLM}-driven {NPC}s: A comparative study of speech recognition and dialogue options,'' in {\em Proceedings of the 30th ACM Symposium on Virtual Reality Software and Technology}, VRST '24, (New York, NY, USA), Association for Computing Machinery, 2024.

\bibitem{155}
F.~Gao, K.~Fang, and W.~K.~V. Chan, ``Humanizing artifacts: An educational game for cultural heritage artifacts and history using generative {AI},'' in {\em Companion Proceedings of the 2024 Annual Symposium on Computer-Human Interaction in Play}, CHI PLAY Companion '24, (New York, NY, USA), p.~91–96, Association for Computing Machinery, 2024.

\bibitem{177}
N.~Jennings, H.~Wang, I.~Li, J.~Smith, and B.~Hartmann, ``What's the game, then? opportunities and challenges for runtime behavior generation,'' in {\em Proceedings of the 37th Annual ACM Symposium on User Interface Software and Technology}, UIST '24, (New York, NY, USA), Association for Computing Machinery, 2024.

\bibitem{150}
A.~Phillips, J.~Lang, and D.~Mould, ``Goal-oriented interactions in games using {LLM}s,'' {\em IEEE Transactions on Games}, pp.~1--12, 2024.

\bibitem{18}
L.~Huang and X.~Sun, ``Create ice cream: Real-time creative element synthesis framework based on gpt3. 0,'' in {\em 2023 IEEE Conference on Games (CoG)}, pp.~1--4, IEEE, 2023.

\bibitem{114}
A.~Isaza-Giraldo, P.~Bala, P.~F. Campos, and L.~Pereira, ``Prompt-gaming: A pilot study on {LLM}-evaluating agent in a meaningful energy game,'' in {\em Extended Abstracts of the 2024 CHI Conference on Human Factors in Computing Systems}, CHI EA '24, (New York, NY, USA), Association for Computing Machinery, 2024.

\bibitem{111}
M.~Pears, C.~Poussa, and S.~T. Konstantinidis, ``Progressive healthcare pedagogy: An application merging {C}hat{GPT} and {AI}-video technologies for gamified and cost-effective scenario-based learning,'' in {\em Interactive Mobile Communication, Technologies and Learning}, pp.~106--113, Springer, 2023.

\bibitem{77}
T.~Shan and K.~Michel, ``Generative {AI} with {GOAP} for fast-paced dynamic decision-making in game environments,'' in {\em 2024 IEEE Conference on Games (CoG)}, pp.~1--5, 2024.

\bibitem{129}
T.~Rist, ``Using a large language model to turn explorations of virtual 3{D}-worlds into interactive narrative experiences,'' in {\em 2024 IEEE Conference on Games (CoG)}, pp.~1--8, 2024.

\bibitem{71}
C.-H. Chen and C.-L. Chang, ``Effectiveness of {AI}-assisted game-based learning on science learning outcomes, intrinsic motivation, cognitive load, and learning behavior,'' {\em Education and Information Technologies}, pp.~1--22, 2024.

\bibitem{96}
A.~Goslen, Y.~J. Kim, J.~Rowe, and J.~Lester, ``{LLM}-based student plan generation for adaptive scaffolding in game-based learning environments,'' {\em International Journal of Artificial Intelligence in Education}, pp.~1--26, 2024.

\bibitem{119}
G.~A. Molina-Barron, R.~E. Alvarado-Ramirez, and A.~Aguirre-Acosta, ``Storytelling: Digital narration enhanced by artificial intelligence in the metaverse,'' in {\em Proceedings of the 2023 7th International Conference on Education and E-Learning}, ICEEL '23, (New York, NY, USA), p.~33–40, Association for Computing Machinery, 2024.

\bibitem{60}
I.~Capecchi, T.~Borghini, and I.~Bernetti, ``{ARF}ood: Pioneering nutrition education for generation alpha through augmented reality and {AI}-driven serious gaming,'' in {\em International Conference on Extended Reality}, pp.~351--359, Springer, 2024.

\bibitem{83}
A.~Rychert, M.~L. Ganuza, and M.~N. Selzer, ``Integrating {GPT} as an assistant for low-cost virtual reality escape-room games,'' {\em IEEE Computer Graphics and Applications}, vol.~44, no.~4, pp.~14--25, 2024.

\bibitem{152}
E.~Whitaker, E.~Trewhitt, and E.~Veinott, ``Heuristica {II}: Updating a 2011 game-based training architecture using generative {AI} tools,'' in {\em International Conference on Human-Computer Interaction}, pp.~314--332, Springer, 2024.

\bibitem{175}
M.~Ribeiro, J.~Santos, J.~a. Lobo, S.~Ara\'{u}jo, L.~Magalh\~{a}es, and T.~Ad\~{a}o, ``{VR}, {AR}, gamification and {AI} towards the next generation of systems supporting cultural heritage: addressing challenges of a museum context,'' in {\em Proceedings of the 29th International ACM Conference on 3D Web Technology}, Web3D '24, (New York, NY, USA), Association for Computing Machinery, 2024.

\bibitem{148}
B.~W. Wilson, S.~Eswaran, O.~Riyaz, J.~Chiyah-Garcia, and M.~P. Aylett, ``Follow the yellow or red brick road? investigating the impact of narratives in a guided navigation task,'' in {\em Proceedings of the 12th International Conference on Human-Agent Interaction}, HAI '24, (New York, NY, USA), p.~411–413, Association for Computing Machinery, 2024.

\bibitem{154}
M.~Sidji, W.~Smith, and M.~J. Rogerson, ``Human-{AI} collaboration in cooperative games: A study of playing codenames with an {LLM} assistant,'' {\em Proc. ACM Hum.-Comput. Interact.}, vol.~8, Oct. 2024.

\bibitem{170}
M.~Shah, M.~Pankiewicz, R.~S. Baker, J.~Chi, Y.~Xin, H.~Shah, and D.~Fonseca, ``Students' use of an {LLM}-powered virtual teaching assistant for recommending educational applications of games,'' in {\em Serious Games} (J.~L. Plass and X.~Ochoa, eds.), (Cham), pp.~19--24, Springer Nature Switzerland, 2025.

\bibitem{143}
L.~Patra, M.~Kumari, and R.~Gopalapillai, ``Enhancing artificial intelligence and machine learning understanding through envision: A virtual reality approach,'' in {\em 2024 First International Conference on Pioneering Developments in Computer Science \& Digital Technologies (IC2SDT)}, pp.~516--521, 2024.

\bibitem{174}
X.~Qin and G.~Weaver, ``Utilizing generative {AI} for {VR} exploration testing: A case study,'' in {\em Proceedings of the 39th IEEE/ACM International Conference on Automated Software Engineering Workshops}, ASEW '24, (New York, NY, USA), p.~228–232, Association for Computing Machinery, 2024.

\bibitem{162}
K.~Plupattanakit, P.~Suntichaikul, P.~Taveekitworachai, R.~Thawonmas, J.~White, K.~Sookhanaphibarn, and W.~Choensawat, ``Llms in eduverse: {LLM}-integrated english educational game in metaverse,'' in {\em 2024 IEEE 13th Global Conference on Consumer Electronics (GCCE)}, pp.~257--258, 2024.

\bibitem{53}
J.~Kelly, M.~Mateas, and N.~Wardrip-Fruin, ``Towards computational support with language models for {TTRPG} game masters,'' in {\em Proceedings of the 18th International Conference on the Foundations of Digital Games}, FDG '23, (New York, NY, USA), Association for Computing Machinery, 2023.

\bibitem{70}
X.~You, P.~Taveekitworachai, S.~Chen, M.~Can~Gursesli, X.~Li, Y.~Xia, and R.~Thawonmas, ``Dungeons, dragons, and emotions: A preliminary study of player sentiment in {LLM}-driven ttrpgs,'' in {\em Proceedings of the 19th International Conference on the Foundations of Digital Games}, FDG '24, (New York, NY, USA), Association for Computing Machinery, 2024.

\bibitem{4}
K.~Frans, ``{AI} charades: Language models as interactive game environments,'' in {\em 2021 IEEE Conference on Games (CoG)}, pp.~1--2, IEEE, 2021.

\bibitem{3}
Y.~Yao, H.~Zhong, Z.~Zhang, X.~Han, X.~Wang, K.~Zhang, C.~Xiao, G.~Zeng, Z.~Liu, and M.~Sun, ``Adversarial language games for advanced natural language intelligence,'' in {\em Proceedings of the AAAI Conference on Artificial Intelligence}, vol.~35, pp.~14248--14256, 2021.

\bibitem{55}
C.~Jaramillo, M.~Charity, R.~Canaan, and J.~Togelius, ``Word autobots: Using transformers for word association in the game codenames,'' {\em Proceedings of the AAAI Conference on Artificial Intelligence and Interactive Digital Entertainment}, vol.~16, pp.~231--237, Oct. 2020.

\bibitem{13}
E.-y. Lee, N.~G.~D. il, G.-h. An, S.~Lee, and K.~Lim, ``{ChatGPT}-based debate game application utilizing prompt engineering,'' in {\em Proceedings of the 2023 International Conference on Research in Adaptive and Convergent Systems}, pp.~1--6, 2023.

\bibitem{42}
T.~S. Wang and A.~S. Gordon, ``Playing story creation games with large language models: Experiments with {GPT}-3.5,'' in {\em Interactive Storytelling} (L.~Holloway-Attaway and J.~T. Murray, eds.), (Cham), pp.~297--305, Springer Nature Switzerland, 2023.

\bibitem{112}
M.~Sidji and M.~Stephenson, ``Prompt engineering {C}hat{GPT} for codenames,'' in {\em 2024 IEEE Conference on Games (CoG)}, pp.~1--4, 2024.

\bibitem{125}
Q.~Ni, Y.~Yu, Y.~Ma, X.~Lin, C.~Deng, T.~Wei, and M.~Xuan, ``The social cognition ability evaluation of llms: A dynamic gamified assessment and hierarchical social learning measurement approach,'' {\em ACM Trans. Intell. Syst. Technol.}, June 2024.
\newblock Just Accepted.

\bibitem{108}
D.~Jeurissen, D.~Perez-Liebana, J.~Gow, D.~Çakmak, and J.~Kwan, ``Playing net{H}ack with {LLM}s: Potential \& limitations as zero-shot agents,'' in {\em 2024 IEEE Conference on Games (CoG)}, pp.~1--8, 2024.

\bibitem{80}
M.~Kim, ``Gpts in mafia-like game simulation,'' in {\em Extended Abstracts of the 2024 CHI Conference on Human Factors in Computing Systems}, CHI EA '24, (New York, NY, USA), Association for Computing Machinery, 2024.

\bibitem{82}
Z.~Shi, M.~Fang, L.~Chen, Y.~Du, and J.~Wang, ``Human-guided moral decision making in text-based games,'' {\em Proceedings of the AAAI Conference on Artificial Intelligence}, vol.~38, pp.~21574--21582, Mar. 2024.

\bibitem{153}
M.~Lamparth, A.~Corso, J.~Ganz, O.~S. Mastro, J.~Schneider, and H.~Trinkunas, ``Human vs. machine: Behavioral differences between expert humans and language models in wargame simulations,'' {\em Proceedings of the AAAI/ACM Conference on AI, Ethics, and Society}, vol.~7, pp.~807--817, Oct. 2024.

\bibitem{161}
S.~Zhong, Z.~Huang, S.~Gao, W.~Wen, L.~Lin, M.~Zitnik, and P.~Zhou, ``Let's think outside the box: Exploring leap-of-thought in large language models with creative humor generation,'' in {\em 2024 IEEE/CVF Conference on Computer Vision and Pattern Recognition (CVPR)}, pp.~13246--13257, 2024.

\bibitem{52}
M.~Ciolino, J.~Kalin, and D.~Noever, ``The {G}o transformer: Natural language modeling for game play,'' in {\em 2020 Third International Conference on Artificial Intelligence for Industries (AI4I)}, pp.~23--26, 2020.

\bibitem{131}
L.~Lawley and C.~Maclellan, ``Val: Interactive task learning with {GPT} dialog parsing,'' in {\em Proceedings of the 2024 CHI Conference on Human Factors in Computing Systems}, CHI '24, (New York, NY, USA), Association for Computing Machinery, 2024.

\bibitem{90}
B.~Bateni and J.~Whitehead, ``Language-driven play: Large language models as game-playing agents in slay the spire,'' in {\em Proceedings of the 19th International Conference on the Foundations of Digital Games}, FDG '24, (New York, NY, USA), Association for Computing Machinery, 2024.

\bibitem{87}
S.~Oh, I.~Chung, and K.-J. Kim, ``Lang{B}irds: An agent for angry birds using a large language model,'' in {\em 2024 IEEE Conference on Games (CoG)}, pp.~1--8, 2024.

\bibitem{139}
A.~L. Davis and G.~Sukthankar, ``Decoding chess mastery: A mechanistic analysis of a chess language transformer model,'' in {\em International Conference on Artificial General Intelligence}, pp.~63--72, Springer, 2024.

\bibitem{160}
Y.~Chen and S.~Chu, ``Large language models in wargaming: Methodology, application, and robustness,'' in {\em 2024 IEEE/CVF Conference on Computer Vision and Pattern Recognition Workshops (CVPRW)}, pp.~2894--2903, 2024.

\bibitem{178}
A.~de~Wynter, ``Will {GPT}-4 run doom?,'' {\em IEEE Transactions on Games}, pp.~1--10, 2024.

\bibitem{84}
Y.~Chen, S.~She, and Y.~Sun, ``Is {AI}-generated content better? a study based on ant forest game content recommendation,'' in {\em Proceedings of the 2024 Guangdong-Hong Kong-Macao Greater Bay Area International Conference on Digital Economy and Artificial Intelligence}, DEAI '24, (New York, NY, USA), p.~749–755, Association for Computing Machinery, 2024.

\bibitem{29}
M.~Lankes and A.~St{\"o}ckl, ``Game reviews reviewed: A game designer’s perspective on {AI}-generated game review analyses,'' in {\em 2023 IEEE Conference on Games (CoG)}, pp.~1--8, IEEE, 2023.

\bibitem{6}
X.~Li, X.~You, S.~Chen, P.~Taveekitworachai, and R.~Thawonmas, ``Analyzing audience comments: Improving interactive narrative with {ChatGPT},'' in {\em International Conference on Interactive Digital Storytelling}, pp.~220--228, Springer, 2023.

\bibitem{171}
U.~B.~R. Khan, N.~Akhtar, A.~Naaz, M.~Akram, and A.~Haque, ``Toward effective suggestion mining in game reviews: Introducing an aspect-based multiclass multilevel dataset,'' in {\em Doctoral Symposium on Computational Intelligence}, pp.~105--119, Springer, 2024.

\bibitem{68}
D.~Carlander, K.~Okada, H.~Engström, and S.~Kurabayashi, ``Controlled chain of thought: Eliciting role-play understanding in {LLM} through prompts,'' in {\em 2024 IEEE Conference on Games (CoG)}, pp.~1--4, 2024.

\bibitem{85}
A.~N. Tak and J.~Gratch, ``Is {GPT} a computational model of emotion?,'' in {\em 2023 11th International Conference on Affective Computing and Intelligent Interaction (ACII)}, pp.~1--8, 2023.

\bibitem{137}
G.~Barbero, M.~M{\"u}ller-Brockhausen, and M.~Preuss, ``Challenges of open world games for {AI}: Insights from human gameplay,'' in {\em Data Science and Artificial Intelligence} (C.~Anutariya, M.~M. Bonsangue, E.~Budhiarti-Nababan, and O.~S. Sitompul, eds.), (Singapore), pp.~127--141, Springer Nature Singapore, 2025.

\bibitem{95}
J.~Liao, S.~Li, Z.~Yang, J.~Wu, Y.~Yuan, X.~Wang, and X.~He, ``L{L}a{RA}: Large language-recommendation assistant,'' in {\em Proceedings of the 47th International ACM SIGIR Conference on Research and Development in Information Retrieval}, SIGIR '24, (New York, NY, USA), p.~1785–1795, Association for Computing Machinery, 2024.

\bibitem{156}
A.~Boz, W.~Zorgdrager, Z.~Kotti, J.~Harte, P.~Louridas, V.~Karakoidas, D.~Jannach, and M.~Fragkoulis, ``Improving sequential recommendations with {LLM}s,'' {\em ACM Trans. Recomm. Syst.}, Jan. 2025.
\newblock Just Accepted.

\bibitem{166}
J.~Zhang, R.~Xie, Y.~Hou, X.~Zhao, L.~Lin, and J.-R. Wen, ``Recommendation as instruction following: A large language model empowered recommendation approach,'' {\em ACM Trans. Inf. Syst.}, Dec. 2024.
\newblock Just Accepted.

\bibitem{113}
P.~Taveekitworachai, F.~Abdullah, M.~C. Gursesli, A.~Lanata, A.~Guazzini, and R.~Thawonmas, ``Prompt evolution through examples for large language models–a case study in game comment toxicity classification,'' in {\em 2024 IEEE International Workshop on Metrology for Industry 4.0 \& IoT (MetroInd4.0 \& IoT)}, pp.~22--27, 2024.

\bibitem{32}
V.~M. Carvalho and M.~A.~F. Rodrigues, ``Investigating and comparing the perceptions of voice interaction in digital games: Opportunities for health and wellness applications,'' in {\em 2023 IEEE 11th International Conference on Serious Games and Applications for Health (SeGAH)}, pp.~1--8, IEEE, 2023.

\bibitem{157}
X.~Liu, A.~Zambrano, A.~Barany, J.~Ocumpaugh, J.~Ginger, M.~Gadbury, H.~C. Lane, and R.~S. Baker, ``Investigating learner interest and observation patterns in a minecraft virtual astronomy environment,'' in {\em International Conference on Quantitative Ethnography}, pp.~19--34, Springer, 2024.

\end{thebibliography}

\bibliographystylereview{ieeetr}

\end{document}